\def\year{2021}\relax
\documentclass[letterpaper]{article} % DO NOT CHANGE THIS
\usepackage{amsmath,amsfonts,bm}

\def\eqref#1{equation~\ref{#1}}
\def\1{\bm{1}}

\def\rvy{{\mathbf{y}}}
\def\rvz{{\mathbf{z}}}

\def\rmA{{\mathbf{A}}}

\def\rmU{{\mathbf{U}}}

\def\rmX{{\mathbf{X}}}

\DeclareMathAlphabet{\mathsfit}{\encodingdefault}{\sfdefault}{m}{sl}
\SetMathAlphabet{\mathsfit}{bold}{\encodingdefault}{\sfdefault}{bx}{n}

\def\sD{{\mathbb{D}}}
\newcommand{\E}{\mathbb{E}}

\newcommand{\KL}{D_{\mathrm{KL}}}

\usepackage{aaai21}  % DO NOT CHANGE THIS
\usepackage{times}  % DO NOT CHANGE THIS
\usepackage{helvet} % DO NOT CHANGE THIS
\usepackage{courier}  % DO NOT CHANGE THIS
\usepackage[hyphens]{url}  % DO NOT CHANGE THIS
\usepackage{graphicx} % DO NOT CHANGE THIS
\usepackage{graphicx}  % DO NOT CHANGE THIS
\usepackage{natbib}  % DO NOT CHANGE THIS AND DO NOT ADD ANY OPTIONS TO IT
\usepackage{caption} % DO NOT CHANGE THIS AND DO NOT ADD ANY OPTIONS TO IT
\usepackage{algorithm}
\usepackage{algorithmic}

\usepackage{amsmath}
\usepackage{amssymb}
\usepackage{booktabs}
\usepackage{multirow}
\usepackage{subfigure}
\usepackage{makecell}
\usepackage{xcolor}
\usepackage[switch]{lineno} 
\begin{document}
\title{SHOT-VAE: Semi-supervised Deep Generative Models \\ With Label-aware ELBO Approximations}
\author {
    % Authors
        Hao-Zhe Feng,\textsuperscript{\rm 1}
        Kezhi Kong,\textsuperscript{\rm 2} 
        Minghao Chen\textsuperscript{\rm 1}\\
        Tianye Zhang,\textsuperscript{\rm 1}
        Minfeng Zhu,\textsuperscript{\rm 1}
        Wei Chen\textsuperscript{\rm 1} \\
}
\affiliations {
    % Affiliations
    \textsuperscript{\rm 1} Zhejiang University \\
    \textsuperscript{\rm 2} University of Maryland, College Park \\
    \{fenghz, zhangtianye1026, chenvis\}@zju.edu.cn, kong@cs.umd.edu,\\
    minghaochen01@gmail.com,  minfeng.zhu@outlook.com
}

% \linenumbers 
\maketitle
\begin{abstract}
Semi-supervised variational autoencoders (VAEs) have obtained strong results, but have also encountered the challenge that \textbf{good} \textbf{ELBO values} \textbf{do not} \textbf{always imply} \textbf{accurate} \textbf{inference} \textbf{results}. In this paper, we investigate and propose two causes of this problem: (1) The ELBO objective cannot utilize the label information directly. (2) A bottleneck value exists, and continuing to optimize ELBO after this value will not improve inference accuracy. On the basis of the experiment results, we propose SHOT-VAE to address these problems without introducing additional prior knowledge. The SHOT-VAE offers two contributions: (1) A new ELBO approximation named \textit{smooth-ELBO} that integrates the label predictive loss into ELBO. (2) An approximation based on \textit{optimal interpolation} that breaks the ELBO value bottleneck by reducing the margin between ELBO and the data likelihood. The SHOT-VAE achieves good performance with 25.30\% error rate on CIFAR-100 with 10k labels and reduces the error rate to 6.11\% on CIFAR-10 with 4k labels.
\end{abstract}
\section{Introduction}
Most deep learning models are trained with large labeled datasets via supervised learning. However, in many scenarios, although acquiring a large amount of original data is easy, obtaining corresponding labels is often costly or even infeasible. Thus, semi-supervised variational autoencoder (VAE) \citep{DBLP:conf/nips/KingmaMRW14} is proposed to address this problem by training classifiers with multiple unlabeled data and a small fraction of labeled data. 

Based on the latent variable assumption \citep{DBLP:journals/corr/Doersch16}, semi-supervised VAE models combine the evidence lower bound (ELBO) and the classification loss as objective, so that it can not only learn the required classification representations from labeled data, but also capture the disentangled factors which could be used for data generation. Although semi-supervised VAE models have obtained strong empirical results on many benchmark datasets (e.g. MNIST, SVHN, Yale B) \citep{DBLP:conf/nips/NarayanaswamyPM17}, it still encounters one common problem that \textbf{good ELBO values do not always imply accurate inference results} \citep{DBLP:journals/corr/ZhaoSE17b}. To address this problem, existing works introduce prior knowledge that needs to be set manually, e.g., the stacked VAE structure (M1+M2, \citealt{DBLP:conf/nips/KingmaMRW14,DBLP:conf/uai/DavidsonFCKT18}), the prior domain knowledge (\citealt{DBLP:journals/corr/LouizosSLWZ15, DBLP:conf/iclr/IlseTLW19}) and mutual information bounds \citep{DBLP:conf/nips/Dupont18}. 

In this study, we investigate the training process of semi-supervised VAE with extensive experiments and propose two possible causes of the problem. (1) First, the ELBO cannot utilize label information directly. In the semi-supervised VAE framework \citep{DBLP:conf/nips/KingmaMRW14}, the classification loss and ELBO learn from the labels and unlabeled data separately, making it difficult to improve the inference accuracy with ELBO. (2) Second, an \textit{``ELBO bottleneck"} exists, and continuing to optimize the ELBO after a certain bottleneck value will not improve inference accuracy. Thus, we propose \textbf{S}moot\textbf{H}-ELBO \textbf{O}ptimal 
In\textbf{T}erpolation VAE (SHOT-VAE) to solve the \textit{``good ELBO, bad performance"} problem without requiring additional prior knowledge, which offers the following contributions:
\begin{itemize}
    \item \textbf{The smooth-ELBO objective that integrates the classification loss into ELBO}. 
    
        We derive a new ELBO approximation named \textit{smooth-ELBO} with the label-smoothing technique \citep{DBLP:conf/nips/MullerKH19}. Theoretically, we prove that the \textit{smooth-ELBO} integrates the classification loss into ELBO. Then, we empirically show that a better inference accuracy can be achieved with \textit{smooth-ELBO}.
    
    \item \textbf{The margin approximation that breaks the ELBO bottleneck.}
    
    We propose an approximation of the margin between the real data distribution and the one from ELBO. The approximation is based on the \textit{optimal interpolation} in data space and latent space. In practice, we show this optimal interpolation approximation (\textit{OT-approximation}) can break the \textit{"ELBO bottleneck"} and achieve a better inference accuracy.
    
    \item \textbf{Good semi-supervised performance.}
    
    We evaluate SHOT-VAE on 4 benchmark datasets and the results show that our model achieves good performance with 25.30\% error rate on CIFAR-100 with 10k labels and reduces the error rate to 6.11\% on CIFAR-10 with 4k labels. Moreover, we find it can get strong results even with fewer labels and smaller models, for example obtaining a 14.27\% error rate on CIFAR-10 with 500 labels and 1.5M parameters.
\end{itemize}
\section{Background}
\label{sec:background}
\subsection{Semi-supervised VAE}
In supervised learning, we are facing with training data that appears as input-label pairs $\displaystyle (\rmX,\rvy)$ sampled from the labeled dataset $\sD_L$. While in semi-supervised learning, we can obtain an extra collection of unlabeled data $\displaystyle \rmX$ denoted by $\sD_U$. We hope to leverage the data from both $\sD_{L}$ and $\sD_{U}$ to achieve a more accurate model than only using $\displaystyle \sD_{L}$. 

Semi-supervised VAEs \cite{DBLP:conf/nips/KingmaMRW14} solve the problem by constructing a probabilistic model to disentangle the data into continuous variables $\rvz$ and label variables $\rvy$. It consists of a generation process and an inference process parameterized by $\bm{\theta}$ and $\bm{\phi}$ respectively. The generation process assumes the posterior distribution of $\rmX$ given the latent variables $\rvz$ and $\rvy$ as 
\begin{equation}
p_{\bm{\theta}}(\rmX\vert \rvz,\rvy)=\mathcal{N}( \rmX;f_{\bm{\theta}}(\rvz,\rvy),\bm{\sigma}^2).
\end{equation}
The inference process assumes the posterior distribution of $\rvz$ and $\rvy$ given $\rmX$ as
\begin{equation}
\begin{split}
     q_{\bm{\phi}}( \rvz \vert \rmX) & = \mathcal{N}(\rvz 
    \vert \bm{\mu}_{\bm{\phi}}(\rmX),\bm{\sigma}^{2}_{\bm{\phi}}(\rmX));\\ 
    q_{\bm{\phi}}( \rvy \vert \rmX) &= \text{Cat}(\rvy\vert \bm{\pi}_{\bm{\phi}}(\rmX)).
\end{split}
\end{equation}
where $\text{Cat}(\rvy\vert \bm{\pi})$ is the multinomial distribution of the label, $\bm{\pi}_{\bm{\phi}}(\rmX)$ is a probability vector, and the functions $f_{\bm{\theta}}$, $\bm{\mu}_{\bm{\phi}}$, $\bm{\sigma}_{\bm{\phi}}$ and $\bm{\pi}_{\bm{\phi}}$ are represented as deep neural networks. 

To make the model learn disentangled representations, the independent assumptions (\citealt{DBLP:conf/nips/KingmaMRW14, DBLP:conf/nips/Dupont18}) are also widely used as 
\begin{equation}
\begin{split}
    p(\rvz,\rvy)&=p(\rvz)p(\rvy);\\
    q_{\bm{\phi}}(\rvz,\rvy\vert \rmX)&=q_{\bm{\phi}}(\rvz\vert \rmX)q_{\bm{\phi}}(\rvy\vert \rmX).
\end{split}
\end{equation}

For the unlabeled dataset $\sD_{U}$, VAE models want to learn the disentangled representation of $q_{\bm{\phi}}( \rvz \vert \rmX)$ and $q_{\bm{\phi}}( \rvy \vert \rmX)$ by maximizing the evidence lower bound of $\log p(\rmX)$ as
\begin{equation}
\begin{split}
        \text{ELBO}_{\sD_{U}}(\rmX) = \E_{q_{\bm{\phi}}(\rvz,\rvy\vert \rmX)} [\log p_{\bm{\theta}}(\rmX\vert \rvz,\rvy)]&\\
        -\KL ( q_{\bm{\phi}}(\rvz\vert \rmX) \Vert p(\rvz)) -\KL ( q_{\bm{\phi}}(\rvy\vert \rmX) \Vert p(\rvy) ) &.
\end{split}
\label{eq:ELBOU}
\end{equation}

For the labeled dataset $\sD_{L}$, the labels $\rvy$ are treated as latent variables and the related ELBO becomes
\begin{equation}
\begin{split}
        \text{ELBO}_{\sD_{L}}(\rmX,\rvy) &= \E_{q_{\bm{\phi}}(\rvz\vert \rmX,\rvy)} [\log p_{\bm{\theta}}(\rmX\vert \rvz,\rvy)]\\
        &-\KL ( q_{\bm{\phi}}(\rvz\vert \rmX,\rvy) \Vert p(\rvz)).
\end{split}
\label{eq:ELBOL}
\end{equation}

Considering the label prediction $q_{\bm{\phi}}(\rvy \vert \rmX)$ contributes only to the unlabeled data in (\ref{eq:ELBOU}), which is an undesirable property as we wish the semi-supervised model can also learn from the given labels, \citet{DBLP:conf/nips/KingmaMRW14} proposes to add a \textit{cross-entropy (CE)} loss as a solution and the extended target is as follows:
\begin{equation}
\begin{split}
        \min_{\bm{\theta},\bm{\phi}} \E_{\rmX \sim \sD_{U}} [-\text{ELBO}&_{\sD_{U}}(\rmX)]+\\ \E_{(\rmX,\rvy) \sim \sD_{L}} [-\text{ELBO}_{\sD_{L}}(\rmX,\rvy) &+\alpha\cdot\text{CE}(q_{\bm{\phi}}(\rvy \vert \rmX),\rvy)]
\end{split}
\label{eq:SSLVAETarget}
\end{equation}
where $\alpha$ is a hyper-parameter controlling the loss weight. 

\subsection{Good ELBO, Bad Inference}
However, a frequent phenomenon is that good ELBO values do not always imply accurate inference results \citep{DBLP:journals/corr/ZhaoSE17b}, which often occurs on realistic datasets with high variance, such as CIFAR-10 and CIFAR-100. In this paper, we investigate the training process of semi-supervised VAE models on the above two datasets and propose two possible causes of the \textit{``good ELBO, bad inference"} problem.

\begin{figure}[h]
\centering
\subfigure[CIFAR-10]{
    \includegraphics[width=1.35in]{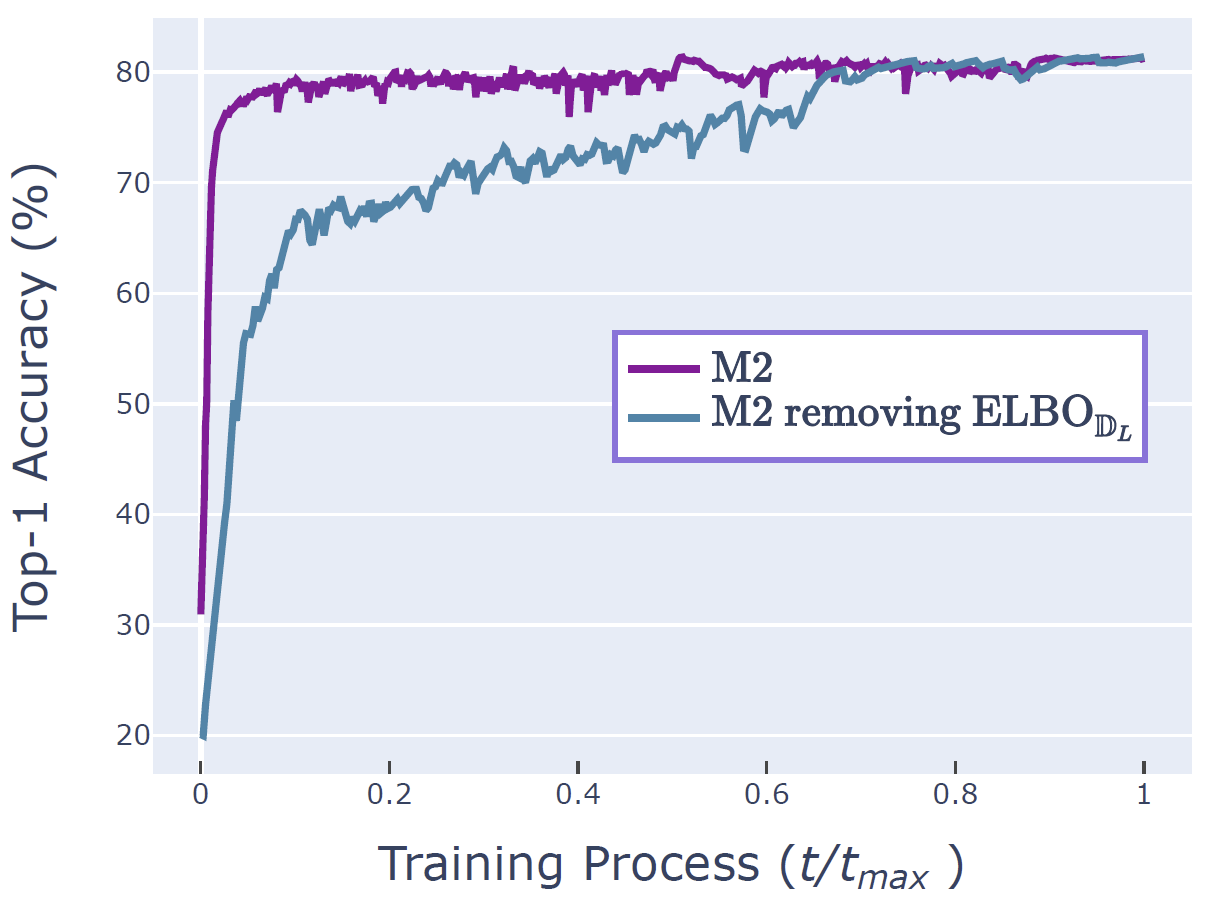}
}
% \quad    %用 \quad 来换行
\subfigure[CIFAR-100]{
\includegraphics[width=1.35in]{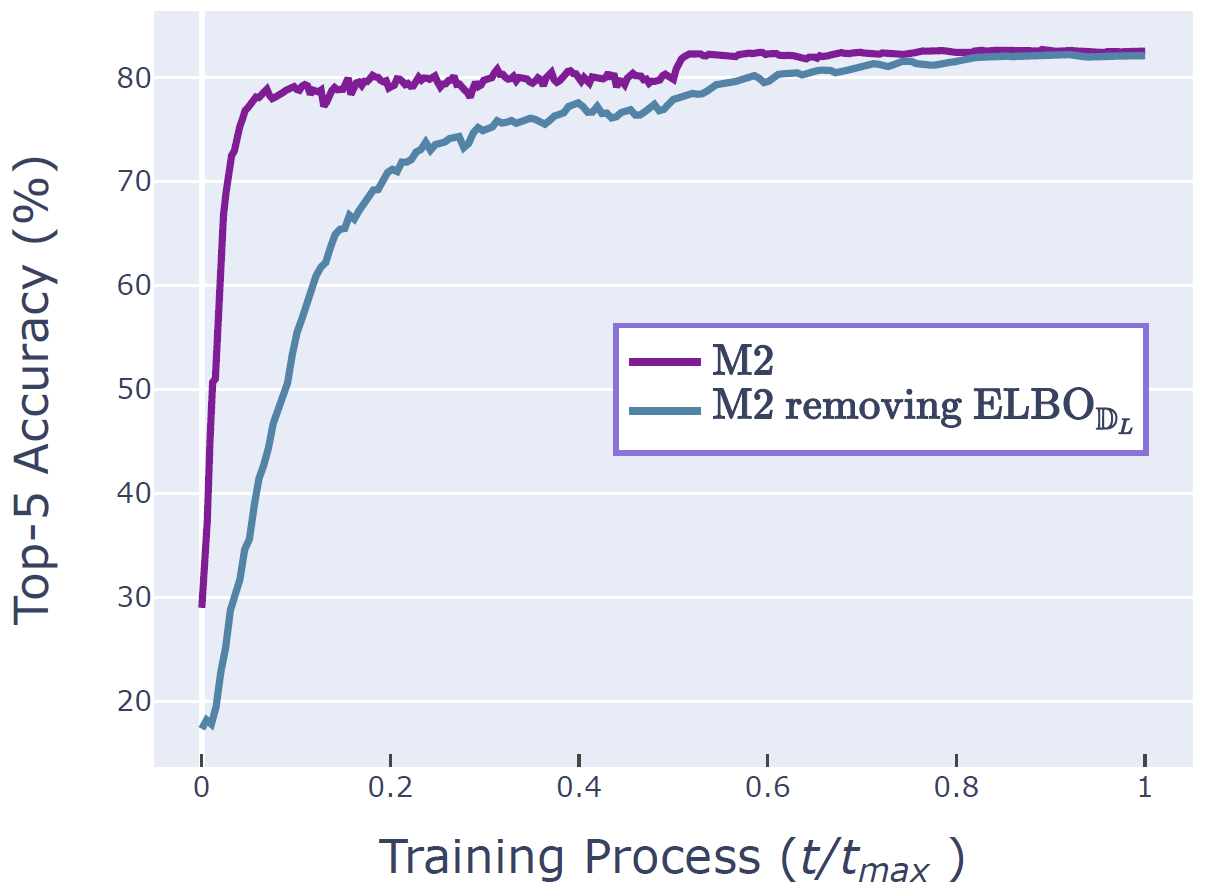}
}
\vspace{-0.15cm}
\caption{Test accuracy of semi-supervised VAE (M2) model with and w/o $\text{ELBO}_{\sD_{L}}$. Results indicates that the $\text{ELBO}_{\sD_{L}}$ fails to achieve a better inference accuracy. }
\vspace{-0.15cm}
\label{fig:frozen-ELBO}
\end{figure}

\textbf{The ELBO cannot utilize the label information.} As mentioned in equation (\ref{eq:SSLVAETarget}), the label prediction $q_{\bm{\phi}}(\rvy \vert \rmX)$ only contributes to the unlabeled loss $-\text{ELBO}_{\sD_{U}}(\rmX)$, which indicates that the labeled loss $-\text{ELBO}_{\sD_{L}}(\rmX,\rvy)$ can not utilize the label information directly. We assume this problem will make the ELBO value irrelevant to the final inference accuracy. To evaluate our assumption, we compare the semi-supervised VAE (M2) models \citep{DBLP:conf/nips/KingmaMRW14} with the same model but removing the $\text{ELBO}_{\sD_{L}}$ in (\ref{eq:SSLVAETarget}). As shown in Figure~\ref{fig:frozen-ELBO}, the results indicate that $\text{ELBO}_{\sD_{L}}$ can accelerate the learning process of $q_{\bm{\phi}}(\rvy \vert \rmX)$, but it fails to achieve a better inference accuracy than the one removing $\text{ELBO}_{\sD_{L}}$. 

\begin{figure}[h]
\centering
\subfigure[CIFAR-10]{
    \includegraphics[width=1.35in]{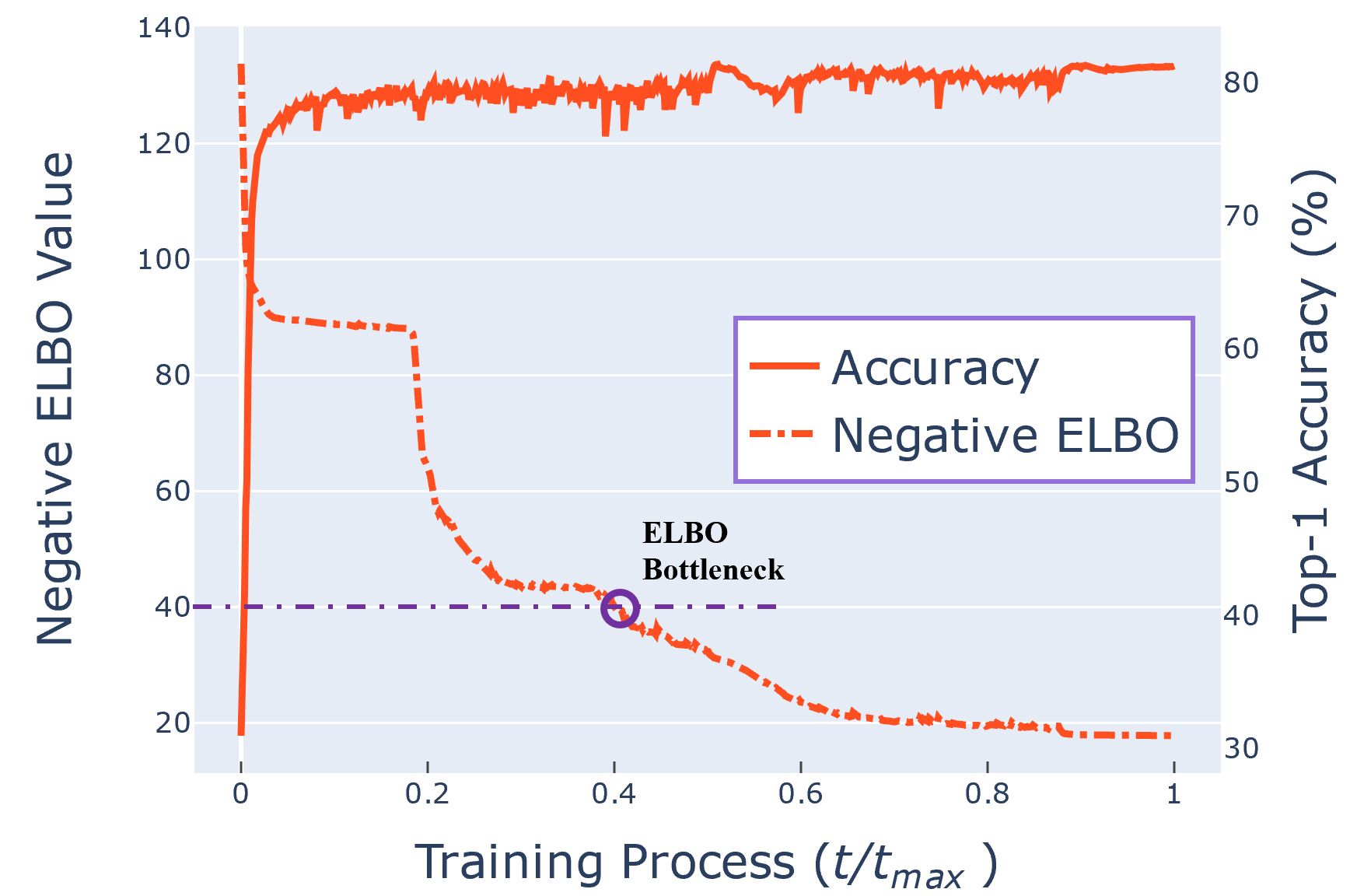}
}
% \quad    %用 \quad 来换行
\subfigure[CIFAR-100]{
\includegraphics[width=1.35in]{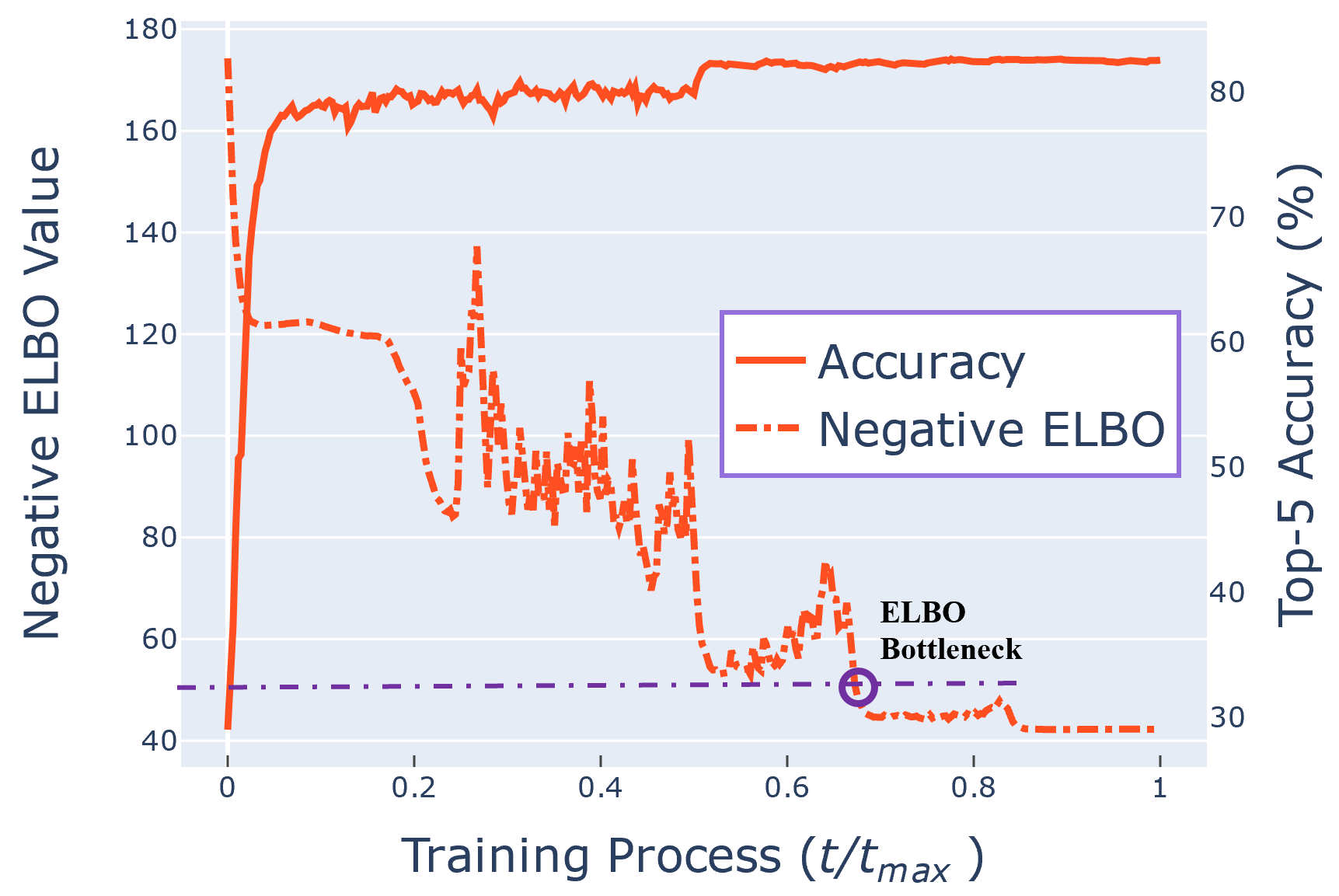}
}
\vspace{-0.15cm}
\caption{Comparison between the negative ELBO value and accuracy for semi-supervised VAE. Results indicate that a ELBO bottleneck exists, and continuing to optimize ELBO after this bottleneck will not improve the inference accuracy.}
\vspace{-0.15cm}
\label{fig:ELBO-Accuracy-Comparison}
\end{figure}

\textbf{The} \textbf{\textit{``ELBO bottleneck"} effect.} Another possible cause is the \textit{``ELBO bottleneck"}, that is, continuing to optimize ELBO after a certain bottleneck value will not improve the inference accuracy. Figure~\ref{fig:ELBO-Accuracy-Comparison} shows that the inference accuracy raises rapidly and peaks at the bottleneck value. After that, the optimization of ELBO value does not affect the inference accuracy. 

Existing works introduce prior knowledge and specific structures to address these problems. \citet{DBLP:conf/nips/KingmaMRW14} and \citet{DBLP:conf/uai/DavidsonFCKT18} propose the stacked VAE structure (M1+M2), which forces the model to utilize the representations learned from $\text{ELBO}_{\sD_{L}}$ to inference the $q_{\bm{\phi}}(\rvy\vert \rmX)$. \citet{DBLP:journals/corr/LouizosSLWZ15} and \citet{DBLP:conf/iclr/IlseTLW19} incorporate domain knowledge into models, making the ELBO representations relevant to the label prediction. \citet{DBLP:journals/corr/ZhaoSE17b} and \citet{DBLP:conf/nips/Dupont18} utilize the ELBO decomposition technique \citep{hoffman2016elbo}, setting the mutual information bounds to perform feature selection. These methods have achieved great success on many benchmark datasets (e.g. MNIST, SVHN, Yale B). However, the related prior knowledge and structures need to be selected manually. Moreover, for some standard datasets with high variance such as CIFAR-10 and CIFAR-100, the semi-supervised performance of VAE is not satisfactory.

Instead of introducing additional prior knowledge, we propose a novel solution based on the ELBO approximations, \textbf{S}moot\textbf{H}-ELBO \textbf{O}ptimal in\textbf{T}erpolation VAE.

\section{SHOT-VAE}
In this section, we derive the SHOT-VAE model by introducing its two improvements. First, for the labeled dataset $\sD_{L}$, we derive a new ELBO approximation named \textit{smooth-ELBO} that unifies the ELBO and the label predictive loss. Then, for the unlabeled dataset $\sD_{U}$, we create the differentiable \textit{OT-approximation} to break the ELBO value bottleneck.
\label{sec:SHOT-VAE}
\subsection{Smooth-ELBO: integrating the classification loss into ELBO.}
\label{subsec:smooth-ELBO-experiment}
To overcome the problem that the $\text{ELBO}_{\sD_{L}}$ cannot utilize the label information directly, we first perform an \textit{``ELBO surgery"}. Following previous works \citep{DBLP:journals/corr/Doersch16}, the $\text{ELBO}_{\sD_{L}}$ can be derived with \textit{Jensen-Inequality} as:
\begin{equation}
    \begin{split}
     & \log p(\rmX,\rvy)=\log\E_{q_{\bm{\phi}}(\rvz\vert \rmX,\rvy)}\frac{p(\rmX,\rvy,\rvz)}{q_{\bm{\phi}}(\rvz\vert \rmX,\rvy)}\\
     & \geq \E_{q_{\bm{\phi}}(\rvz\vert \rmX,\rvy)}\log\frac{p(\rmX,\rvy,\rvz)}{q_{\bm{\phi}}(\rvz\vert \rmX,\rvy)}=\text{ELBO}_{\sD_{L}}(\rmX,\rvy)
    \end{split}
    \label{eq:elbo-surgery-1}
\end{equation}

Utilizing the independent assumptions in equation $(3)$, we have $q_{\bm{\phi}}(\rvz\vert \rmX,\rvy)=q_{\bm{\phi}}(\rvz\vert \rmX)$. In addition, the labels $\rvy$ are treated as latent variables directly in $\text{ELBO}_{\sD_{L}}$, which equals to obey the empirical degenerate distribution, i.e. $\hat{p}(\rvy_{i}\vert \rmX_{i})=1,\forall (\rmX_i,\rvy_i)\in \sD_{L}$. Substituting the above two conditions into (\ref{eq:elbo-surgery-1}), we have
\begin{equation}
    \begin{split}
       \text{ELBO}_{\sD_{L}}&(\rmX,\rvy) =  \E_{
       q_{\bm{\phi}}(\rvz\vert \rmX),\hat{p}(\rvy\vert\rmX)}\log\frac{p(\rmX,\rvy,\rvz)}{q_{\bm{\phi}}(\rvz\vert \rmX)\hat{p}(\rvy\vert\rmX)}\\
       &=\E_{
       q_{\bm{\phi}},\hat{p}}\log p(\rmX\vert \rvz,\rvy)-\KL(q_{\bm{\phi}}(\rvz\vert \rmX)\Vert p(\rvz))\\
       &-\KL(\hat{p}(\rvy\vert\rmX)\Vert p(\rvy)).
    \end{split}
    \label{eq:elbo-surgery-2}
\end{equation}
In equation (\ref{eq:elbo-surgery-2}), the last objective $\KL(\hat{p}(\rvy\vert \rmX) \Vert p(\rvy))$ is irrelevant to the label prediction $q_{\bm{\phi}}(\rvy \vert \rmX)$, which causes the \textit{“good ELBO, bad inference”} problem. Inspired by this, we derive a new ELBO approximation named \textit{smooth-ELBO}. 

The \textit{smooth-ELBO} provides two improvements. First, we propose a more flexible assumption of the empirical distribution $\hat{p}(\rvy\vert \rmX)$. Instead of treating $\hat{p}(\rvy\vert \rmX)$ as the degenerate distribution, we use the label smoothing technique \citep{DBLP:conf/nips/MullerKH19} and view the one-hot label $\1_{\rvy}$ as the parameters of the empirical distribution $\hat{p}(\rvy \vert \rmX)$, that is, $\forall (\rmX,\rvy) \in \sD_{L}$
\begin{equation}
\begin{split}
    \hat{p}(\rvy \vert \rmX) &= \text{Cat}(\rvy \vert \text{smooth}(\1_{\rvy}));\\ \text{smooth}(\1_{\rvy})_{i} &= \begin{cases}1-\epsilon&\text{if } \1_{\rvy,i}=1,\\
         \frac{\epsilon}{K-1}&\text{if } \1_{\rvy,i}=0.\end{cases},
\end{split}
\label{eq:smooth-label-1}
\end{equation}
where $K$ is the number of classes and $\epsilon$ controls the smooth level. We use $\epsilon=0.001$ in all experiments. 

Then, we derive the following convergent approximation with the smoothed $\hat{p}(\rvy\vert \rmX)$:
\begin{equation}
    \begin{split}
      \KL(\hat{p}(\rvy\vert \rmX) \Vert& q_{\bm{\phi}}(\rvy\vert \rmX))
    +\KL ( q_{\bm{\phi}}(\rvy\vert \rmX) \Vert p(\rvy) )\\
    &\rightarrow \KL(\hat{p}(\rvy\vert \rmX) \Vert p(\rvy)) \\
        &\text{when } q_{\bm{\phi}}(\rvy\vert \rmX) \rightarrow \hat{p}(\rvy\vert \rmX)  .
    \end{split}
    \label{eq:smooth-elbo-1}
\end{equation}

The proof can be found in Appendix A and we also point out the approximation (\ref{eq:smooth-elbo-1}) does not converge under the empirical degenerate distribution, which explains the importance of label smoothing. Combining equations (\ref{eq:elbo-surgery-2}), (\ref{eq:smooth-label-1}) and (\ref{eq:smooth-elbo-1}), we propose the \textit{smooth-ELBO} objective for $\sD_{L}$:
\begin{equation}
\begin{split}
    \text{smooth-E}&\text{LBO}_{\sD_{L}}(\rmX,\rvy) \\
    = \E_{q_{\bm{\phi}},\hat{p}}\log p(\rmX\vert \rvz,\rvy)&-
    \KL(q_{\bm{\phi}}(\rvz\vert \rmX)\Vert p(\rvz))\\
    -\KL ( q_{\bm{\phi}}(\rvy\vert \rmX) \Vert p(\rvy) )
    &-\KL(\hat{p}(\rvy\vert \rmX) \Vert q_{\bm{\phi}}(\rvy\vert \rmX)).
\end{split}
\end{equation}
Theoretically, we demonstrate the following properties.

\textbf{Smooth-ELBO integrates the classification loss into ELBO.} Compared with the original $\text{ELBO}_{\sD_{L}}$, \textit{smooth-ELBO} derives two extra components, $\KL ( q_{\bm{\phi}}(\rvy\vert \rmX) \Vert p(\rvy) )$ and $\KL(\hat{p}(\rvy\vert \rmX) \Vert q_{\bm{\phi}}(\rvy\vert \rmX))$. Utilizing the decomposition in \citep{hoffman2016elbo}, we can rewrite the first component into 
\[
\displaystyle
\E_{\sD_{L}}\KL ( q_{\bm{\phi}}(\rvy\vert \rmX) \Vert p(\rvy) ) = \mathbf{I}_{q_{\bm{\phi}}}(\rmX;\rvy) + \KL ( q_{\bm{\phi}}(\rvy) \Vert p(\rvy) )
\]
where $\mathbf{I}_{q_{\bm{\phi}}}(\rmX;\rvy)$ is the constant of mutual information between $\rmX$ and $\rvy$, $p(\rvy)$ is the true marginal distribution for $\rvy$ which can be estimated with $\hat{p}(\rvy\vert \rmX)$ in $\sD{_L}$ and $q_{\bm{\phi}}(\rvy) = \frac{1}{\vert \sD_{L} \vert} \sum_{(\rmX,\rvy)\in \sD_{L}}q_{\bm{\phi}}(\rvy\vert \rmX)$ is the estimation of marginal distribution. By optimizing $ \KL ( q_{\bm{\phi}}(\rvy) \Vert p(\rvy) )$, the first component can learn the marginal distribution $p(\rvy)$ from labels. 

For the second component, with Pinsker's inequality, it's easy to prove that for all $i=1,2,\ldots, K$ 
\[
    \begin{split}
        \vert \bm{\pi}_{\bm{\phi}}(\rmX)-\text{smooth}(\1_{\rvy})\vert_{i}
        \leq \sqrt{\frac{1}{2} \KL(\hat{p}(\rvy\vert \rmX) \Vert q_{\bm{\phi}}(\rvy\vert \rmX))}
    \end{split}
\]
The proof can be found in Appendix B, which indicates that $q_{\bm{\phi}}(\rvy\vert \rmX)$ converges to $\hat{p}(\rvy\vert \rmX)$ in training process.

\textbf{Convergence analysis.} As mentioned above, $q_{\bm{\phi}}(\rvy\vert \rmX)$ converge to the smoothed $\hat{p}(\rvy\vert \rmX)$ in training. Based on this property, we can assert that the \textit{smooth-ELBO} converges to $\text{ELBO}_{\sD_{L}}$ with the following equation:
\begin{equation*}
 \vert \text{smooth-ELBO}_{\sD_{L}}(\rmX,\rvy)-\text{ELBO}_{\sD_{L}}(\rmX,\rvy)\vert \leq C_1\delta+C_2\frac{\delta^2}{\epsilon}+\Delta(\delta)
\end{equation*}
The proof can be found in Appendix C. $C_1,C_2$ are the constants related to class number $K $ and $\delta=\sup_{i}\vert \bm{\pi}_{\bm{\phi}}(\rmX)_{i}-\text{smooth}(\1_{\rvy})_{i}\vert$ is the distance between $q_{\bm{\phi}}(\rvy\vert \rmX)$ and $\hat{p}(\rvy\vert \rmX)$.

To summarize the above, \textbf{smooth-ELBO can utilize the label information directly}. Compared with the original $-\text{ELBO}_{\sD_{L}}+\alpha \text{CE}$ loss in (\ref{eq:SSLVAETarget}), \textit{smooth-ELBO} has three advantages. First, it not only learns from single labels, but also learns from the marginal distribution $p(\rvy)$. Second, we do not need to manually set the loss weight $\alpha$. Moreover, it also takes advantages of the $\text{ELBO}_{\sD_{L}}$, such as disentangled representations and convergence assurance. The extensive experiments will also show that a better model performance can be achieved with smooth-ELBO.

\subsection{OT-approximation: breaking the ELBO bottleneck}
\label{subsec:optimal-interpolation-approximation}
To overcome the \textit{ELBO bottleneck} problem, we first analyze what the semi-supervised VAE model does after reaching the bottleneck, then we create the differentiable \textit{OT-approximation} to break it, which is based on the \textit{optimal interpolation} in latent space. 

As mentioned in equation (\ref{eq:ELBOU}), VAE aims to learn disentangled representations $q_{\bm{\phi}}(\rvz\vert \rmX)$ and $q_{\bm{\phi}}(\rvy\vert \rmX)$ by maximizing the lower bound of the likelihood of data as $\log p(\rmX)\geq \text{ELBO}(\rmX)$, while the margin between $\log p(\rmX)$ and $\text{ELBO}(\rmX)$ has the following closed form:
\begin{equation}
\begin{split}
        \log p(\rmX)-\text{ELBO}(\rmX) = \KL(q_{\bm{\phi}}(\rvz\vert \rmX)q_{\bm{\phi}}(\rvy\vert \rmX) \Vert p(\rvz,\rvy\vert \rmX)) 
\end{split}
\label{eq:ELBO-closed-form}
\end{equation}

Ideally, the optimization process of ELBO will make the representation $q_{\bm{\phi}}(\rvz\vert \rmX)$ and $q_{\bm{\phi}}(\rvy\vert \rmX)$ converge to their ground truth $p(\rvz\vert \rmX)$ and $p(\rvy\vert \rmX)$. However, the unimproved inference accuracy of $q_{\bm{\phi}}(\rvy\vert \rmX)$ indicates that \textbf{optimizing ELBO after the bottleneck will only contribute to the continuous representation $q_{\bm{\phi}}(\rvz\vert \rmX)$, while the $q_{\bm{\phi}}(\rvy\vert \rmX)$ seems to get stuck in the local minimum}. Since the ground truth $p(\rvy\vert \rmX)$ is not available for the unlabeled dataset $\sD_{U}$, it is hard for the model to jump out by itself. Inspired by this, we create a differentiable approximation of  $\KL(q_{\bm{\phi}}(\rvy\vert \rmX) \Vert p(\rvy\vert \rmX))$ for $\sD_{U}$ to break the bottleneck. 

Following previous works \citep{lee2013pseudo}, the approximation is usually constructed with two steps: creating the pseudo input $\Tilde{\rmX}$ with data augmentations and creating the pseudo distribution $\Tilde{p}(\rvz\vert \Tilde{\rmX})$ of $\Tilde{\rmX}$. Recent advanced works use autoaugment \citep{Cubuk_2019_CVPR} and random mixmatch \citep{DBLP:journals/corr/abs-1905-02249} to perform data augmentations. However, these strategies will greatly change the representation of continuous variable $\rvz$, e.g., changing the image style and background. To overcome this, we propose the \textit{optimal interpolation} based approximation.  

The optimal interpolation consists of two steps. First, for each input $\rmX_0$ in $\sD_{U}$, we find the optimal match $\rmX_1$ with the most similar continuous variable $\rvz$, that is,  $\arg_{\rmX_1 \in \sD_{U}}\min\KL(q_{\bm{\phi}}(\rvz\vert \rmX_0) \Vert q_{\bm{\phi}}(\rvz\vert \rmX_1))$. Then, on purpose of jumping out the stuck point $q_{\bm{\phi}}(\rvy\vert \rmX)$, we take the widely-used mixup strategy \citep{DBLP:conf/iclr/ZhangCDL18} to create pseudo input $\Tilde{\rmX}$ as follows:
\begin{equation}
\begin{split}
    \Tilde{\rmX}=(1-\lambda)\rmX_{0}+\lambda\rmX_{1},
\end{split}
\label{eq:mixup}
\end{equation}
where $\lambda$ is sampled from the uniform distribution $\rmU(0,1)$. 

The mixup strategy can be understood as calculating the \text{optimal interpolation} between two points $\rmX_{0},\rmX_{1}$ in input space with the maximum likelihood:
\begin{equation*}
\begin{split}
        \max_{\Tilde{X}} (1-\lambda)\cdot \log(p_{\bm{\theta}}(\Tilde{\rmX}\vert \rvz_0,\rvy_0))
        +\lambda\cdot \log(p_{\bm{\theta}}(\Tilde{\rmX}\vert \rvz_1,\rvy_1)),
\end{split}
\end{equation*}
where $\{\rvz_{i},\rvy_{i}\}_{i=0,1}$ is the latent variables for the data points $\rmX_{0},\rmX_{1}$, and the proof can be found in Appendix D.
\begin{algorithm}[t] 
\caption{SHOT-VAE training process with epoch $t$.} 
\label{alg:SHOT-VAE} 
\begin{algorithmic}[1] 
\REQUIRE ~~\\ %算法的输入参数：Input
Batch of labeled data $(\rmX_{L},\rvy_{L})\in\sD_{L}$;\\ Batch of unlabeled data $\rmX_{U}\in \sD_{U}$;\\
Mixup $\lambda\sim \rmU(0,1)$;\\
Loss weight $w_t$;\\
Model parameters: $\bm{\theta}^{(t-1)},\bm{\phi}^{(t-1)}$;\\ 
Model optimizer: $\text{SGD}$
\ENSURE ~~\\ %算法的输出：Output
Updated parameters: $\bm{\theta}^{(t)},\bm{\phi}^{(t)}$
\STATE $L_{\sD_{L}}=- \text{smooth-ELBO}_{\sD_{L}}(\rmX_{L},\rvy_{L})$
\STATE $\rmX_{U}^{0},\rmX_{U}^{1}=\rmX_{U},\text{OptimalMatch}(\rmX_{U})$
\STATE $L_{\sD_{U}}=    -\text{ELBO}_{\sD_{U}}(\rmX_{U})+w_t\cdot \text{OT}_{\sD_{U}}(\rmX_{U}^{0},\rmX_{U}^{1},\lambda)$
\STATE $L=L_{\sD_{L}}+L_{\sD_{U}}$
\STATE $\bm{\theta}^{(t)},\bm{\phi}^{(t)}=\text{SGD}(\bm{\theta}^{(t-1)},\bm{\phi}^{(t-1)},\frac{\partial L}{\partial \bm{\theta}},\frac{\partial L}{\partial \bm{\phi}})$
\end{algorithmic}
\end{algorithm}

To create the pseudo distribution $\Tilde{p}(\rvy\vert \Tilde{\rmX})$ of $\Tilde{\rmX}$, it is a natural thought that \textbf{the optimal interpolation in data space could associate with the same in latent space} with $\KL$ distance used in ELBO. Inspired by this, we propose the optimal interpolation method to calculate $\Tilde{p}(\rvy\vert \Tilde{\rmX})$ as

\textbf{Proposition 1} \textit{The optimal interpolation derived from $\KL$ distance between $q_{\bm{\phi}}(\rvy\vert \bm{\pi}_{\bm{\phi}}(\rmX_0))$ and $q_{\bm{\phi}}(\rvy\vert \bm{\pi}_{\bm{\phi}}(\rmX_1))$  with $\lambda \in [0,1]$ can be written as}
\begin{equation*}
\begin{split}
        \min_{\Tilde{\bm{\pi}}} (1-\lambda)\cdot\KL(\bm{\pi}_{\bm{\phi}}(\rmX_0)\Vert \Tilde{\bm{\pi}})+\lambda\cdot \KL(\bm{\pi}_{\bm{\phi}}(\rmX_1)\Vert \Tilde{\bm{\pi}})
\end{split}
\end{equation*}
\textit{and the solution $\Tilde{\bm{\pi}}$ satisfying}
\begin{equation}
    \Tilde{\bm{\pi}} = (1-\lambda) \bm{\pi}_{\bm{\phi}}(\rmX_0) + \lambda\bm{\pi}_{\bm{\phi}}(\rmX_1).
\end{equation}
The proof can be found in Appendix E. 

Combining the optimal interpolation in data space and latent space, we derive the \textbf{o}ptimal in\textbf{t}erpolation approximation (OT-approximation) for $\sD_{U}$ as
\begin{equation}
\begin{split}
        &\text{OT}_{\sD_{U}}(\rmX_0,\rmX_1,\lambda)=\KL(q_{\bm{\phi}}(\rvy\vert \Tilde{\rmX}) \Vert \Tilde{p}(\rvy\vert \Tilde{\rmX}))\\
        &\textbf{s.t. }\begin{cases}
\Tilde{\rmX}=(1-\lambda)\rmX_{0}+\lambda\rmX_{1}\\
\Tilde{p}(\rvy\vert \Tilde{\rmX})= \text{Cat}(\rvy\vert \Tilde{\bm{\pi}})\\
\Tilde{\bm{\pi}} = (1-\lambda)\bm{\pi}_{\bm{\phi}}(\rmX_0)+\lambda\bm{\pi}_{\bm{\phi}}(\rmX_1)
\end{cases}.
\end{split}
\end{equation}

Notice that the \textit{OT-approximation} does not require additional prior knowledge and is easy to implement. Moreover, although \textit{OT-approximation} utilizes the mixup strategy to create pseudo input $\Tilde{\rmX}$, our work has two main advantages over mixup-based methods (\citealt{DBLP:conf/iclr/ZhangCDL18,DBLP:conf/icml/VermaLBNMLB19}). First, mixup methods directly assume the pseudo label $\Tilde{\rvy}$ behaves linear in latent space without explanations. Instead, we derive the $\mathcal{D}_{KL}$ from ELBO as the metric and utilize the optimal interpolation $(15)$ to construct $\Tilde{\rvy}$. Second, mixup methods use $\Vert\cdot\Vert_2^2$ loss between $q_{\bm{\phi}}(\rvy\vert \Tilde{\rmX})$ and $\Tilde{p}(\rvy\vert \Tilde{\rmX}))$, while we use the $\mathcal{D}_{KL}$ loss and achieve better semi-supervised learning performance.
\begin{figure*}[t]
\centering
\subfigure{
    \includegraphics[width=3.2in]{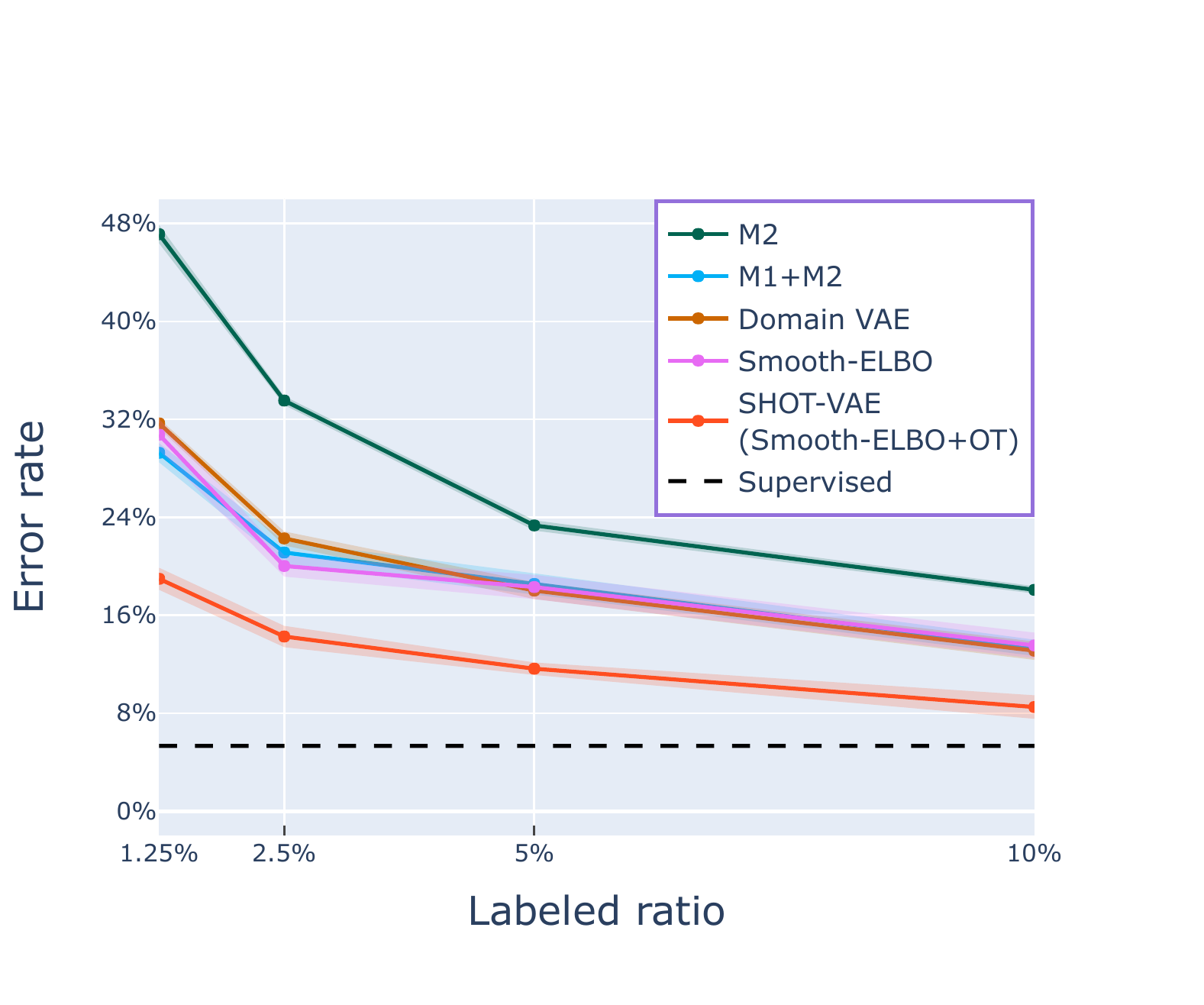}
}   %用 \quad 来换行
\subfigure{
\includegraphics[width=3.2 in]{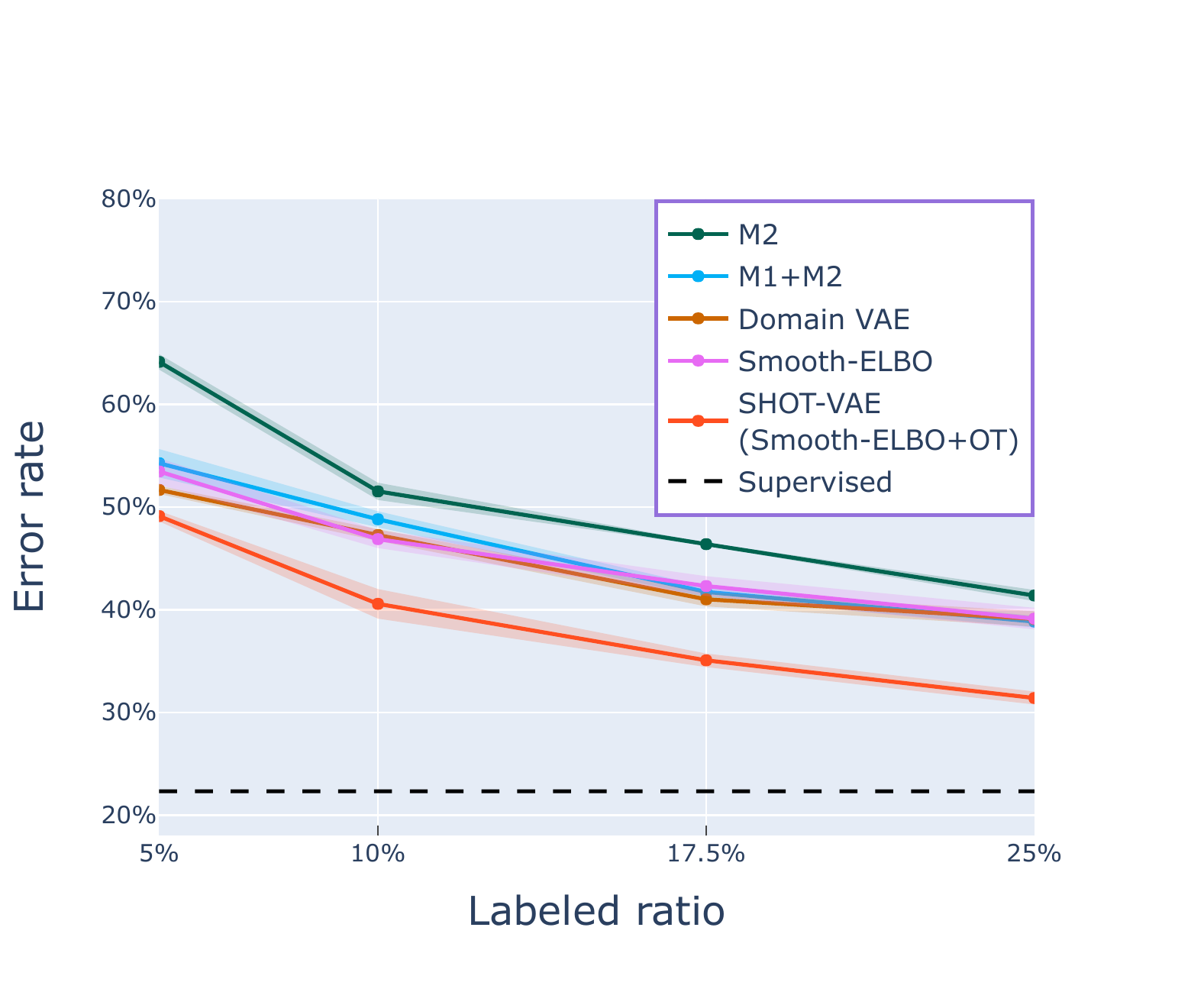}
}
\vspace{-0.2in}
\caption{Error rate comparison of SHOT-VAE to baseline methods on CIFAR-10 (left) and CIFAR-100 (right) for a varying number of labels. ``Supervised" refers to training with all 50000 training samples and no unlabeled data. Results show that (1) SHOT-VAE surpasses other models with a large margin in all cases. (2) both \textit{smooth-ELBO} and \textit{OT-approximation} contribute to the inference accuracy, reducing the error rate on $10\%$ labels from $18.08\%$ to $13.54\%$ and from $13.54\%$ to $8.51\%$.}
\label{fig:vary-ratios-label}
\end{figure*}
\subsection{The implementation details of SHOT-VAE}
The complete algorithm of SHOT-VAE can be obtained by combining the \textit{smooth-ELBO} and the \textit{OT-approximation}, as shown in Algorithm \ref{alg:SHOT-VAE}. In this section, we discuss some details in training process.

First, the working condition for \textit{OT-approximation} is that the ELBO has reached the bottleneck value. However, quantifying the ELBO bottleneck value is difficult. Therefore, we extend the \textit{\textit{warm-up strategy}} in $\beta-\textbf{VAE}$ \citep{DBLP:conf/iclr/HigginsMPBGBML17} to achieve the working condition. The main idea of \textit{\textit{warm-up}} is to make the weight $w_t$ for OT-approximation increase slowly at the beginning and most rapidly in the middle of the training, i.e. exponential schedule. The function of the exponential schedule is
\begin{equation}
    w_t = \exp(-\gamma\cdot(1-\frac{t}{t_{\text{max}}})^2),
\end{equation}
where $\gamma$ is the hyper-parameter controlling the increasing speed, and we use $\gamma=5$ in all experiments.

Second, the optimal match operation in equation $(13)$ requires to find the most similar $\rmX_1$ for each $\rmX_0$ in $\sD_{U}$, which consumes a lot of computation resources. To overcome this, we set a large batch-size (e.g., 512) and use the most similar $\rmX_1$ in one mini-batch to perform optimal interpolation.

Moreover, calculating the gradient of the expected log-likelihood $\E_{q_{\bm{\phi}}(\rvz,\rvy\vert \rmX)} \log p_{\bm{\theta}}(\rmX\vert \rvz,\rvy)$ is difficult. Therefore, we apply the reparameterization tricks  \cite{DBLP:conf/icml/RezendeMW14,DBLP:conf/iclr/JangGP17} to obtain the gradients as follows:
\begin{equation*}
    \begin{split}
        &\nabla_{\bm{\theta},\bm{\phi}}\E_{q_{\bm{\phi}}(\rvz\vert \rmX)}\log p_{\bm{\theta}}(\rmX\vert \rvz) \approx(\bm{\epsilon}_i\sim\mathcal{N}(\mathbf{0},\mathbf{I}))\\
        &\frac{1}{N}\sum_{i=1}^{N}\nabla_{\bm{\theta},\bm{\phi}}\log p_{\bm{\theta}}(\rmX\vert\bm{\mu}_{\bm{\phi}}(\rmX)+\bm{\sigma}_{\bm{\phi}}(\rmX) \odot \bm{\epsilon}_i ).
    \end{split}
\end{equation*}
and
\begin{equation*}
    \begin{split}
        &\nabla_{\bm{\theta},\bm{\phi}}\E_{q_{\bm{\phi}}(\rvy\vert \rmX)}\log p_{\bm{\theta}}(\rmX\vert \rvy)\approx(\bm{\epsilon}_i\sim\textbf{Gumbel}(\bm{\epsilon};\mathbf{0},\mathbf{1}))\\ 
        &\frac{1}{N}\sum_{i=1}^{N}\nabla_{\bm{\theta},\bm{\phi}}\log p_{\bm{\theta}}(\rmX\vert \text{Softmax}(\frac{\log \bm{\pi}_{\bm{\phi}}(\rmX)+\bm{\epsilon}_i}{\tau})).
    \end{split}
\end{equation*}
Following previous works \citep{DBLP:conf/nips/Dupont18}, we used $N=1$ and $\tau=0.67$ in all experiments.

To make the VAE model learn the disentangled representations, we also take the widely-used $\beta$-VAE strategy \citep{DBLP:journals/corr/abs-1804-03599} in training process and chose $\beta=0.01$ in all experiments.

\section{Experiments}
\label{sec:experiments}
In this section, we evaluate the SHOT-VAE model with sufficient experiments on four benchmark datasets, i.e. MNIST, SVHN, CIFAR-10, and CIFAR-100. In all experiments, we apply stochastic gradient descent (SGD) as optimizer with momentum $0.9$ and multiply the learning rate by $0.1$ at regularly scheduled epochs. For each experiment, we create five $\sD_{L}$-$\sD_{U}$ splits with different random seeds and the error rates are reported by the mean and variance across splits. Due to space limitations, we mainly show results on CIFAR-10 and CIFAR-100; more results on MNIST and SVHN as well as the robustness analysis of hyper-parameters are provided in Appendix F. The code, with which the most important results can be reproduced, is available at Github\footnote{\url{https://github.com/PaperCodeSubmission/AAAI2021-260}}. 

\begin{table*}[t]
\vskip -0.2cm
\centering
\begin{tabular}{ccccc}
\toprule
\makecell[c]{Parameter\\Amount}   & Method & 
\makecell[c]{CIFAR10\\(4k)}&
\makecell[c]{CIFAR100\\(4k)} & \makecell[c]{CIFAR100\\(10k)}\\
\midrule
\multirow{5}{*}{1.5 M}
&VAT&$13.13$& /\ & $37.78$ \\
&$\Pi$-Model&$16.37$& /\ &$39.19$\\
&Mean Teacher&$15.87$& $44.71$ & $38.92$ \\
&CT-GAN& $10.62$ &$45.11$ & $37.16$\\
&LP&$11.82$& $43.73$ & $35.92$\\
&Mixup& $10.71(\pm 0.44)$&$46.61(\pm 0.88)$ &$38.62(\pm 0.67)$\\
&\bf{SHOT-VAE} & $\mathbf{8.51}(\pm 0.32)$& $\mathbf{40.58}(\pm 0.48)$ &$\mathbf{31.41}(\pm 0.21)$ \\
\midrule
\multirow{4}{*}{36.5 M}
& $\Pi$-Model& $12.16$ & /\ &$31.12$ \\
&Mean Teacher & $6.28$& $36.63$ & $27.71$ \\
&MixMatch& $\mathbf{5.53}$ &$35.62$&$25.88$\\
&\bf{SHOT-VAE} & $6.11(\pm 0.34)$ & $\mathbf{33.76}(\pm 0.53)$& $\mathbf{25.30}(\pm 0.34)$\\
\bottomrule
\end{tabular}
\caption{Error rate comparison of SHOT-VAE to baseline models on CIFAR-10 and CIFAR-100 with 4k and 10k labels in different parameter amounts. The results show that SHOT-VAE outperforms other advanced methods on CIFAR-100. Moreover, our model is not sensitive to the parameter amount. For example, the accuracy on CIFAR-10 only loses 2\% when the model size decreases 24 times.}
\label{table:SHOT-VAE-CIFAR}
\end{table*}
\begin{table}[t]
\vspace{-0.3cm}
\centering
\begin{tabular}{ccc}
\toprule
$t/t_{max}$ & CIFAR-10 & CIFAR-100 \\
\midrule
0.1         &   $0.71(\pm 0.24)\%$       & $2.79(\pm 0.6)\%$          \\
0.5         &   $0.39(\pm 0.13)\%$       & $1.61(\pm 0.57)\%$       \\
1           &   $0.11(\pm 0.04)\%$       & $0.46(\pm 0.11)\%$     \\
\bottomrule
\end{tabular}
\caption{Relative error of smooth-ELBO.}
\label{table:convergence speed}
\vspace{-0.4cm}
\end{table}
\subsection{Smooth-ELBO improves the inference accuracy}
In the above sections, We propose \textit{smooth-ELBO} as the alternative of $-\text{ELBO}_{\sD_{L}}$+ $\text{CE}$ loss in equation (\ref{eq:SSLVAETarget}), and analyze the convergence theoretically. Here we evaluate the inference accuracy and convergence speed of \textit{smooth-ELBO}. 

First, we compare the \textit{smooth-ELBO} with other semi-supervised VAE models under a varying label ratios from $1.25\%$ to $25\%$. As baselines, we consider three advanced VAE models mentioned above: standard semi-supervised VAE (M2), stacked-VAE (M1+M2), and domain-VAE \citep{DBLP:conf/iclr/IlseTLW19}. 

As shown in Figure \ref{fig:vary-ratios-label}, \textit{smooth-ELBO} makes VAE model learn better representations from labels, reducing the error rates among all label ratios on CIFAR-10 and CIFAR-100 respectively. Moreover, \textit{smooth-ELBO} also achieves competitive results to other VAE models without introducing additional domain knowledge or multi-stage structures.

Second, we analyze the convergence speed in training process. As mentioned above, the \textit{smooth-ELBO} will converge to the real ELBO when $q_{\bm{\phi}}(\rvy\vert \rmX) \rightarrow \hat{p}(\rvy\vert \rmX)$. Moreover, we also descover that $q_{\bm{\phi}}(\rvy\vert \rmX)$ converges to $\hat{p}(\rvy\vert \rmX)$ in training process. Here we evaluate the convergence speed in training process with the relative error between the \textit{smooth-ELBO} and the real ELBO. As shown in Table \ref{table:convergence speed}, the relative error can be very low even at the early stage of training, that is, $0.71\%$ on CIFAR-10 and $2.79\%$ on CIFAR-100, which indicates that the \textit{smooth-ELBO} converges rapidly.

\subsection{SHOT-VAE breaks the ELBO bottleneck}
\label{subsec:SHOT-VAE}
In the above sections, we make two assertions: (1) optimizing ELBO after the bottleneck will make $q_{\bm{\phi}}(\rvy\vert \rmX)$ get stuck in the local minimum. (2) The OT-approximation can break the ELBO bottleneck by making good estimation of $\KL(q_{\bm{\phi}}(\rvy\vert \rmX) \Vert p(\rvy\vert \rmX))$ for $\rmX \in \sD_{U}$. Here we evaluate the assertions through two stage experiments. 

\begin{figure}[t]
\vskip -0.5cm
\centering
\subfigure[CIFAR-10]{\includegraphics[width=1.5in]{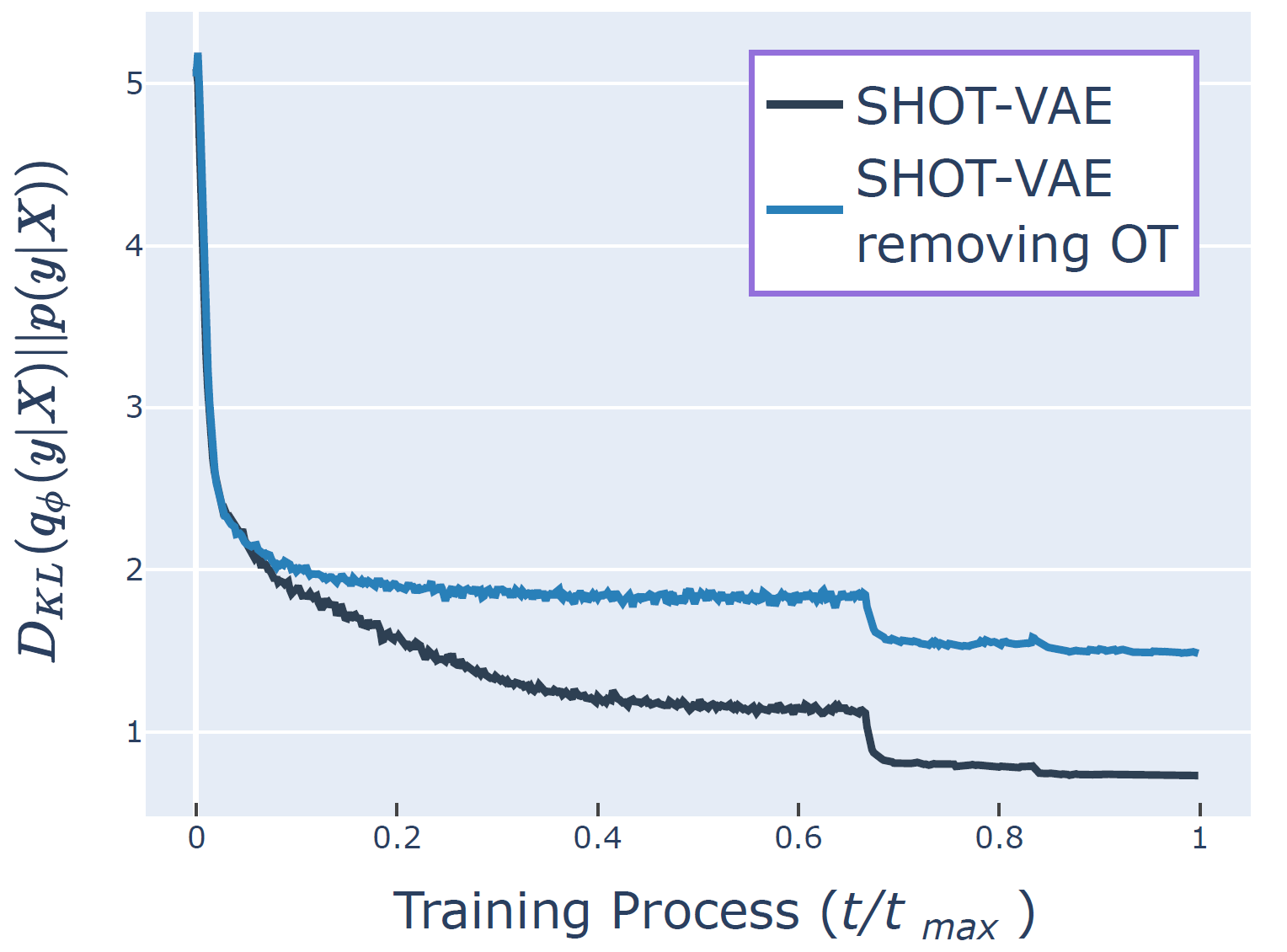}}
% \quad    %用 \quad 来换行
\subfigure[CIFAR-100]{
\includegraphics[width=1.5in]{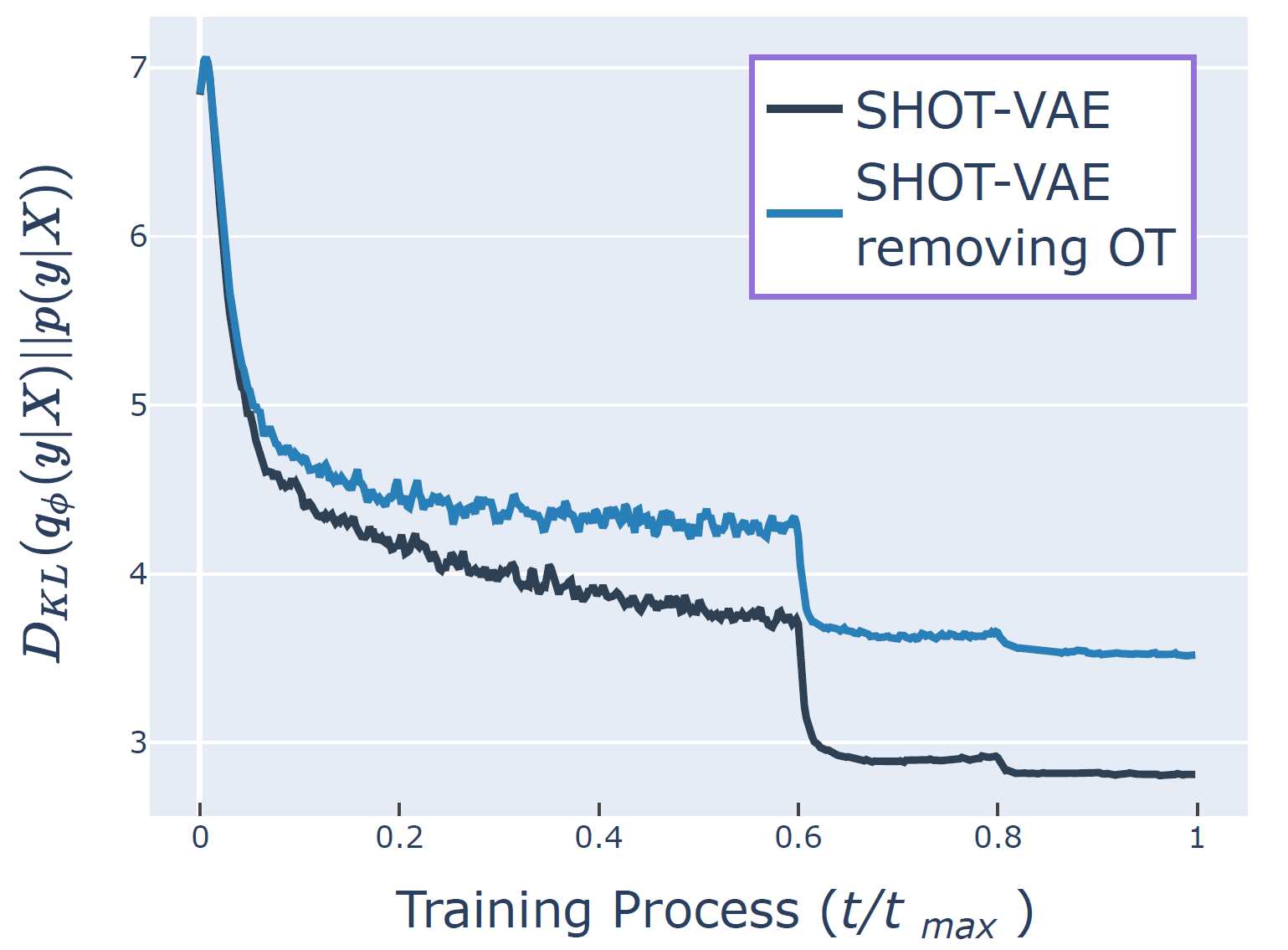}
}
\vspace{-0.1in}
\caption{The $\KL(q_{\bm{\phi}}(\rvy\vert \rmX) \Vert \hat{p}(\rvy\vert \rmX))$ in $\sD_{U}$ with and w/o \textit{OT-approximation}. Results indicate that the OT approximation bridges the gap between  $q_{\bm{\phi}}(\rvy\vert \rmX)$ and $\hat{p}(\rvy\vert \rmX)$ in $\sD_{U}$, making $q_{\bm{\phi}}(\rvy\vert \rmX)$ jump out the local minimum.}
\label{fig:ot-bridge-the-gap}
\vskip -0.35cm
\end{figure}

First, to evaluate the ``local minimum" assertion, we utilize the label of $\sD_{U}$ to estimate the empirical distribution $\hat{p}(\rvy\vert \rmX)$ and calculate $\KL(q_{\bm{\phi}}(\rvy\vert \rmX) \Vert \hat{p}(\rvy\vert \rmX))$ in training process as the metric. Notice these labels are only used to calculate the metric and do not contribute to the model. As shown in Figure \ref{fig:ot-bridge-the-gap}, we compare the SHOT-VAE with the same model but removing the \textit{OT-approximation}. The results indicate that optimizing ELBO itself without \textit{OT-approximation} will make the gap $\KL(q_{\bm{\phi}}(\rvy\vert \rmX) \Vert \hat{p}(\rvy\vert \rmX))$ stuck into the local minimum, while the \textit{OT approximation} helps the model jump out the local minimum, leading to a better inference of $q_{\bm{\phi}}(\rvy\vert \rmX)$.

\begin{figure}[ht]
    \centering
\subfigure[CIFAR-10]{
    \includegraphics[width=1.5in]{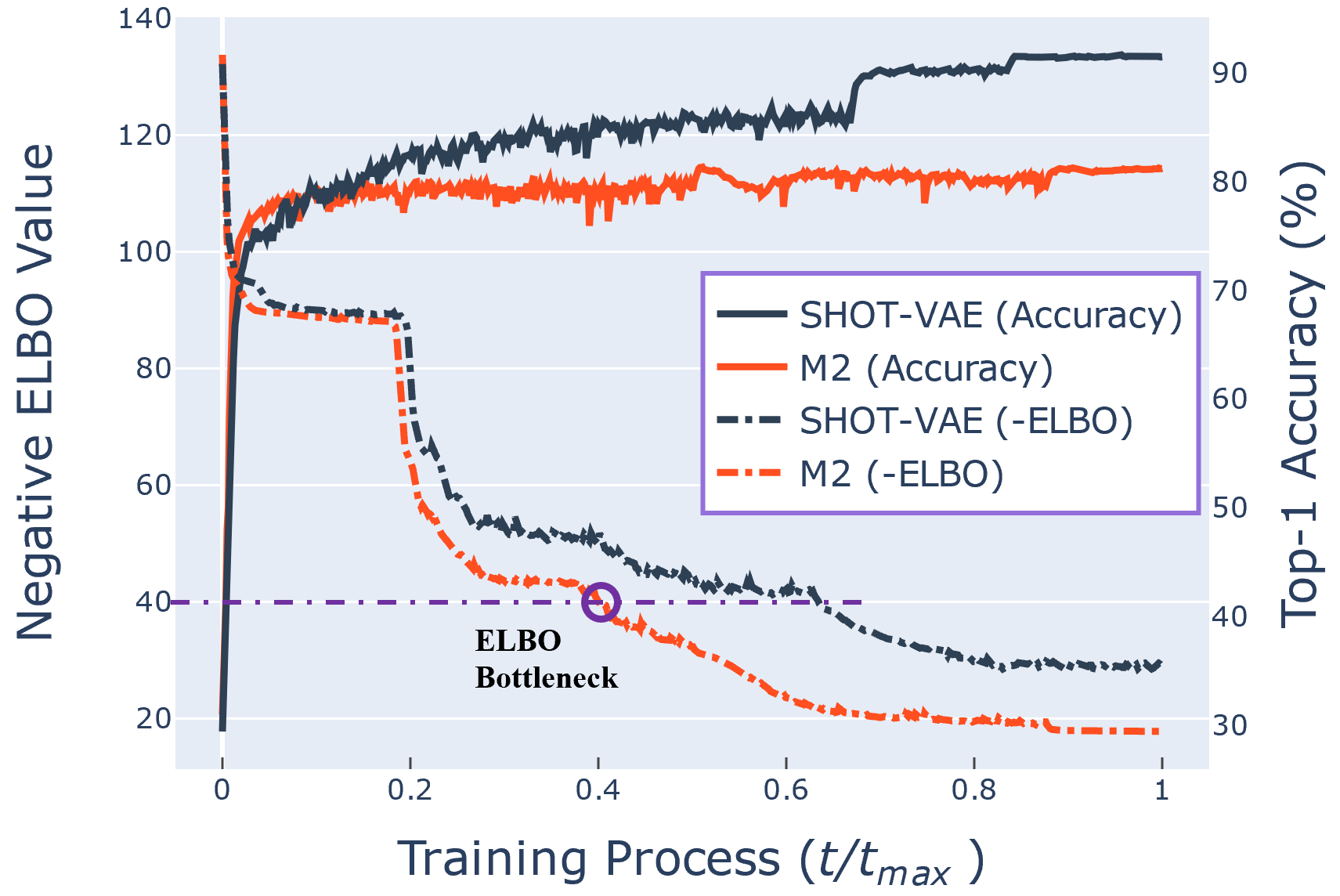}
}
% \quad    %用 \quad 来换行
\subfigure[CIFAR-100]{
\includegraphics[width=1.5in]{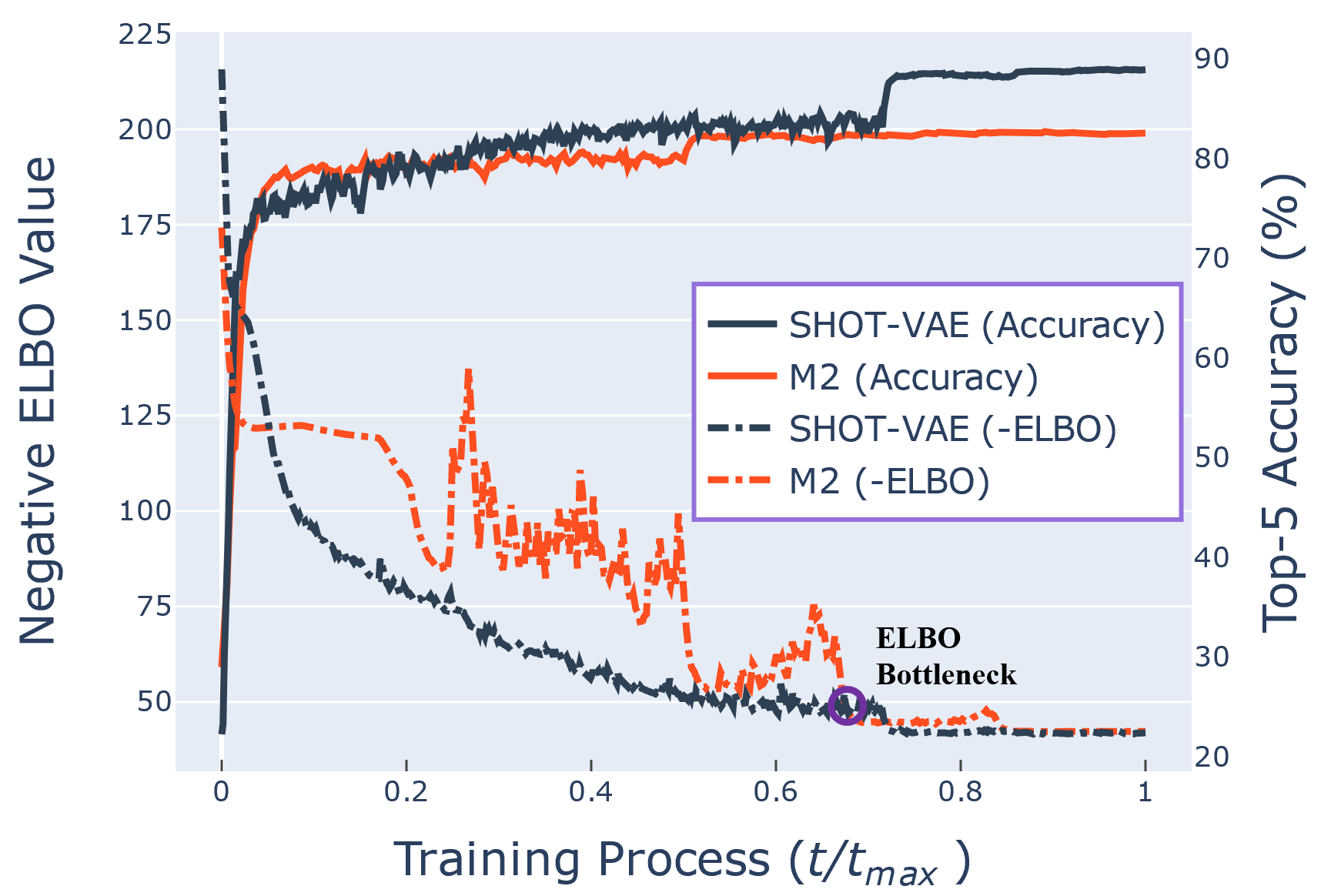}
}
\vspace{-0.2in}
\caption{Comparison between the negative ELBO value and test accuracy for SHOT-VAE and M2 model. Results indicate that SHOT-VAE breaks the ELBO bottleneck.}
\label{fig:break-ELBO}
\end{figure}

Then, we investigate the relation between the negative ELBO and inference accuracy for SHOT-VAE and M2 model. As shown in Figure \ref{fig:break-ELBO}, for M2 model, the inference accuracy stalls after the ELBO bottleneck. While for SHOT-VAE, optimizing ELBO contributes to the improvement of inference accuracy during the whole training process. Moreover, the SHOT-VAE achieves a much better accuracy than M2, which also indicates that SHOT-VAE breaks the ``\textit{ELBO bottleneck}".
\subsection{Semi-supervised learning performance}
\label{subsec:semi-perform}
We evaluate the effectiveness of the SHOT-VAE on two parts: evaluations under a varying number of labeled samples and evaluations under different parameter amounts of neural networks. 

First, we compare the SHOT-VAE with other advanced VAE models under a varying label ratios from $1.25\%$ to $25\%$. As shown in Figure \ref{fig:vary-ratios-label}, both \textit{smooth-ELBO} and \textit{OT-approximation} contribute to the improvement of inference accuracy, reducing the error rate on $10\%$ labels from $18.08\%$ to $13.54\%$ and from $13.54\%$ to $8.51\%$, respectively. Furthermore, SHOT-VAE outperforms all other methods by a large margin, e.g., reaching an error rate of $14.27\%$ on CIFAR-10 with the label ratio $2.5\%$. For reference, with the same backbone, fully supervised training on all 50000 samples achieves an error rate of $5.33\%$.

Then, we evaluate SHOT-VAE under different parameter amounts of neural networks, i.e. ``WideResNet-28-2" with 1.5M parameters and ``WideResNet-28-10" with 36.5M parameters. As baselines for comparison, we select six current best models from 4 categories: Virtual Adversarial Training (VAT \citet{DBLP:journals/pami/MiyatoMKI19}) and MixMatch \citep{DBLP:journals/corr/abs-1905-02249} which are based on data augmentation, $\Pi$-model \citep{DBLP:conf/iclr/LaineA17} and Mean Teacher \citep{DBLP:conf/nips/TarvainenV17}, based on model consistency regularization, Label Propagation (LP) \citep{iscen2019label} based on pseudo-label and CT-GAN \citep{DBLP:conf/iclr/WeiGL0W18} based on generative models. The results are presented in Table \ref{table:SHOT-VAE-CIFAR}. Besides, we also take the mixup-based method into consideration. In general, the SHOT-VAE model outperforms other methods among all experiments on CIFAR-100. Furthermore, our model is not sensitive to the parameter amount and reaches competitive results even with small networks (e.g., 1.5M parameters).

Moreover, our SHOT-VAE can easily combine other advanced semi-supervised methods, which further improves model performance. As shown in Appendix G, we combine SHOT-VAE with data augmentations and mean teacher separately and achieve much better results, i.e. $4.56\%$ error rate on CIFAR-10 (4k) and $24.09\%$ on CIFAR-100 (10k). 
\subsection{Disentangled representations}
\begin{figure}[t]
\vspace{-0.06in}
\centering
\subfigure[MNIST]{
    \includegraphics[width=2.8in]{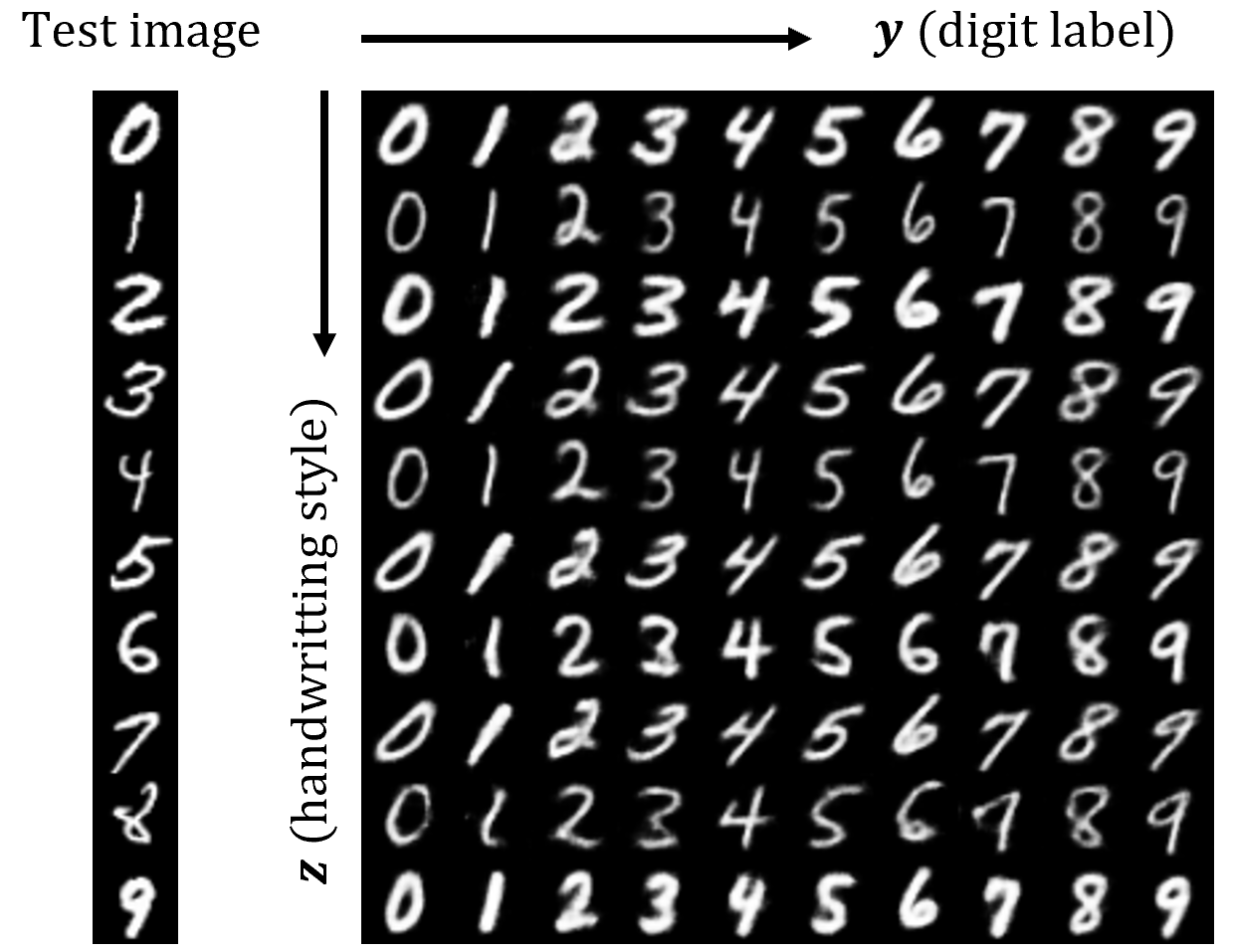}
}
\quad    %用 \quad 来换行
\subfigure[SVHN]{
\includegraphics[width=2.8in]{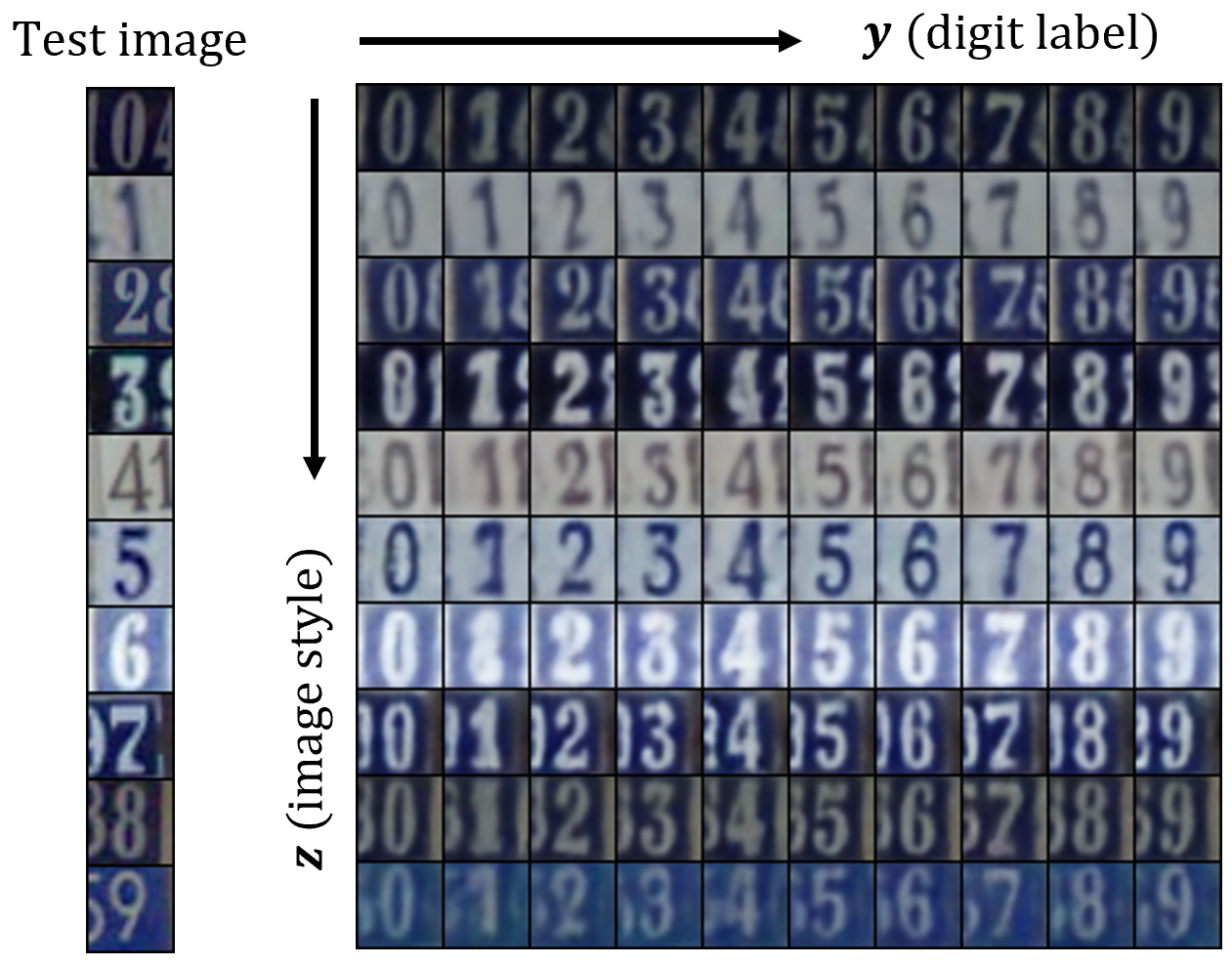}
}
\vspace{-0.1in}
\caption{The conditional generation results of SHOT-VAE. The leftmost columns show images from the test set and the other columns show the conditional generation samples with the learned representation. It indicates that $\rvz$ and $\rvy$ have learned disentangled representations in latent space, as $\rvz$ represents the image style and $\rvy$ represents the digit label. }
\vspace{-0.1in}
\label{fig:conditional-generation}
\end{figure}

Among semi-supervised models, VAE based approaches have great advantages in interpretability by capturing semantics-disentangled latent variables. To demonstrate this property, we perform conditional generation experiments on MNIST and SVHN datasets. As shown in Figure \ref{fig:conditional-generation}, we pass the test image through the inference network to obtain the distribution of the latent variables $\rvz$ and $\rvy$ corresponding to this image. We then fix the inference $q_{\bm{\phi}}(\rvz \vert \rmX)$ of continuous variable $\rvz$, vary $\rvy$ with different labels, and generate new samples. The generation results show that $\rvz$ and $\rvy$ have learned semantic-disentangled representations, as $\rvz$ represents the image style and $\rvy$ represents the classification contents. Moreover, by comparing the results through columns, we find that each dimension of the discrete variable $\rvy$ corresponds to one class label separately.

\section{Conclusions}
We investigate one challenge in semi-supervised VAEs that ``good ELBO values do not imply accurate inference results". We propose two causes of this problem through reasonable experiments. Based on the experiment results, we propose SHOT-VAE to address the \textit{``good ELBO, bad inference"} problem. With extensive experiments, We demonstrate that SHOT-VAE can break the ELBO value bottleneck without introducing additional prior knowledge. Results also show that our SHOT-VAE outperforms other advanced semi-supervised models. 
\clearpage
\section{Ethic statement}
We address the SHOT-VAE to solve one common problem in semi-supervised generative models that good ELBO values do not always imply accurate inference results. We offer the possible influence of our work from three aspects: fundamental theory impact, machine learning application impact, and social impact.

For the fundamental theory impact, we propose two causes of the \textit{``good ELBO, bad inference"} problem through extensive experiments. Based on the experiment results, we provide two contributions: (1) A new ELBO approximation named smooth-ELBO that integrates the label predictive loss into ELBO. (2) An approximation based on optimal interpolation that breaks the ELBO value bottleneck. The two approximations are all reasonable in theory and work in practice. 

For the machine learning application impact. We evaluate the SHOT-VAE under realistic datasets with high variance and show it has great advantages in interpretability by capturing semantics-disentangled latent variables.

For the social impact, the SHOT-VAE can achieve good performance with a small fraction of labeled data. Therefore, it is friendly to the data providers and the governments that pay attention to the privacy-preserving policy, e.g., the general data protection regulation in the European Union.

\bibliography{aaai21}
\section{Appendix}
\setcounter{equation}{0}
\subsection{Appendix A}
\label{Appendix.A}
\textbf{Proposition 1}  \textit{The following limitations hold for the smoothed empirical distribution $\hat{p}(\rvy\vert \rmX)$ as defined in the Section \textit{smooth-ELBO}, when $q_{\bm{\phi}}(\rvy\vert \rmX) \rightarrow \hat{p}(\rvy\vert \rmX) $:}
\begin{equation}
    \begin{split}
      \KL(\hat{p}(\rvy\vert \rmX) \Vert& q_{\bm{\phi}}(\rvy\vert \rmX))
    +\KL ( q_{\bm{\phi}}(\rvy\vert \rmX) \Vert p(\rvy) )\\
    &\rightarrow \KL(\hat{p}(\rvy\vert \rmX) \Vert p(\rvy)) \\
    \end{split}
    \label{eq:smooth-elbo-1}
\end{equation}
\textbf{proof :}

To analyze the convergence of $(1)$, we first decompose the $\KL(\hat{p}(\rvy\vert \rmX) \Vert p(\rvy))$ as follows

\begin{equation}
\begin{split}
        &\KL(\hat{p}(\rvy\vert \rmX) \Vert p(\rvy)) =\E_{\hat{p}(\rvy\vert \rmX)}\log\frac{\hat{p}(\rvy\vert \rmX)}{p(\rvy)}\\
        &=\E_{\hat{p}(\rvy\vert \rmX)}\log\frac{\hat{p}(\rvy\vert \rmX)}{q_{\bm{\phi}}(\rvy\vert \rmX)}+\E_{\hat{p}(\rvy\vert \rmX)}\log\frac{q_{\bm{\phi}}(\rvy\vert \rmX)}{p(\rvy)}\\
        &=\KL(\hat{p}(\rvy\vert \rmX) \Vert q_{\bm{\phi}}(\rvy\vert \rmX))+\E_{\hat{p}(\rvy\vert \rmX)}\log\frac{q_{\bm{\phi}}(\rvy\vert \rmX)}{p(\rvy)}
\end{split}
\label{eq:decomposition-1}
\end{equation}
Since $\KL ( q_{\bm{\phi}}(\rvy\vert \rmX) \Vert p(\rvy) )=-\E_{q_{\bm{\phi}}(\rvy\vert \rmX)}\log \frac{p(\rvy)}{q_{\bm{\phi}}(\rvy\vert \rmX)}$, the proof for (\ref{eq:smooth-elbo-1}) is equals to prove the following limitation:
\begin{equation*}
\lim_{q_{\bm{\phi}}(\rvy\vert \rmX) \rightarrow \hat{p}(\rvy\vert \rmX)  }\E_{q_{\bm{\phi}}(\rvy\vert \rmX)}\log \frac{p(\rvy)}{q_{\bm{\phi}}(\rvy\vert \rmX)}=\E_{\hat{p}(\rvy\vert \rmX)}\log \frac{p(\rvy)}{q_{\bm{\phi}}(\rvy\vert \rmX)}.
\end{equation*}
The proof is as follows. 

First, we explicitly define the condition $q_{\bm{\phi}}(\rvy\vert \rmX) \rightarrow \hat{p}(\rvy\vert \rmX)$ with a closed form. That is, $\forall \zeta \geq 0$, there exists at least one $\delta$ satisfies 
\begin{equation}
\begin{split}
    \E_{\hat{p}(\rvy\vert \rmX)}\log \frac{p(\rvy)}{q_{\bm{\phi}}(\rvy\vert \rmX)} -\E_{q_{\bm{\phi}}(\rvy\vert \rmX)}\log \frac{p(\rvy)}{q_{\bm{\phi}}(\rvy\vert \rmX)}&\leq \zeta\\
       \text{when } \sup_{i}  \vert \bm{\pi}_{\bm{\phi}}(\rmX)_{i}-\text{smooth}(\1_{\rvy})_{i}\vert &\leq \delta
\end{split}
    \label{eq:limitation}
\end{equation}
\begin{figure}[ht]
\vspace{-0.5cm}
\centering
\includegraphics[width=2.8in]{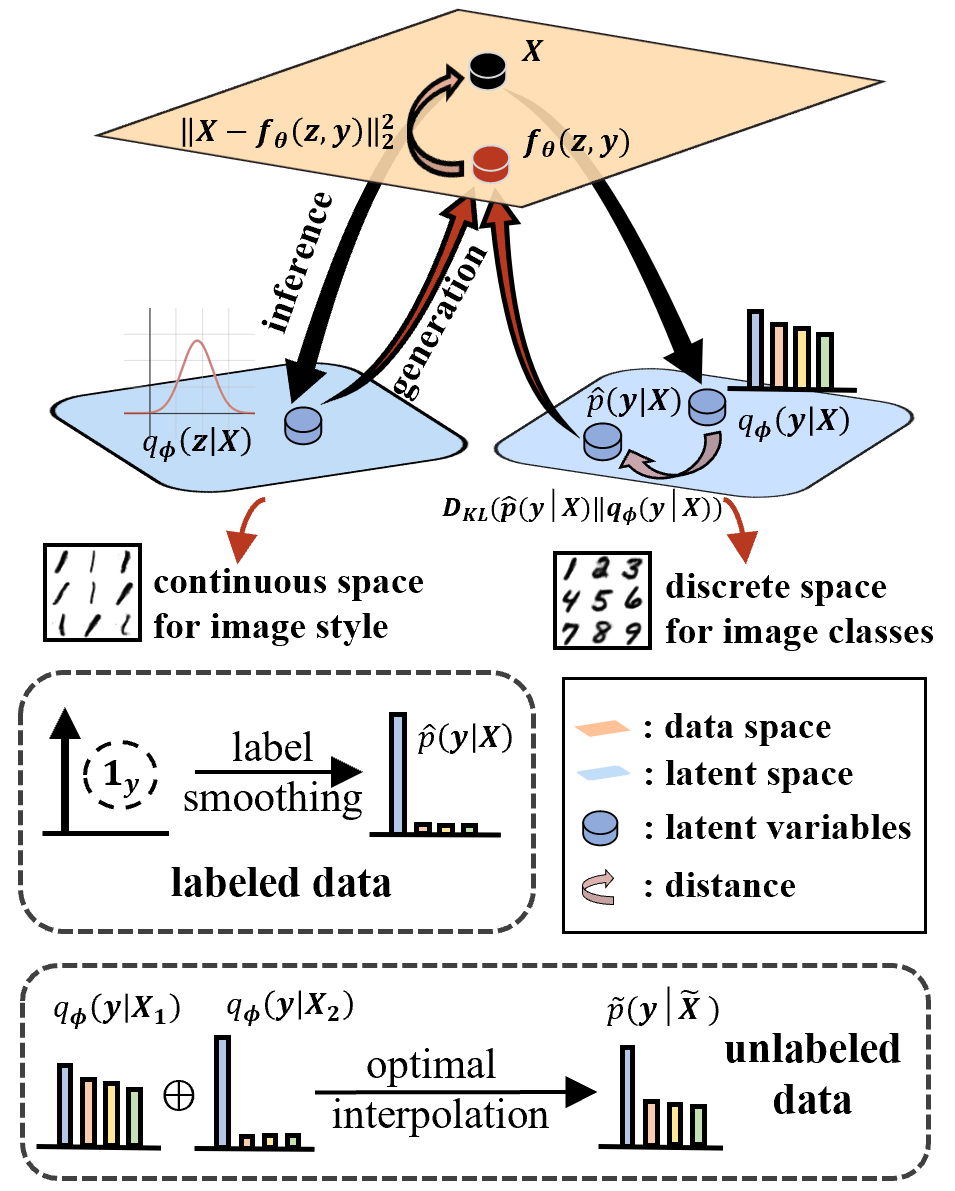}
\vspace{-0.2cm}
\caption{The schematic of SHOT-VAE. SHOT-VAE has great advantages in interpretability by capturing semantics-disentangled latent variables as $\rvz$ represents the image style and $\rvy$ represents the image class. The smooth-ELBO proposes a more flexible assumption of $\hat{p}(\rvy\vert \rmX)$ with label-smoothing technique and the optimal interpolation performs data augmentation on the input pairs with the most similar continuous representations and breaks the ELBO bottleneck.}
\vspace{-0.5cm}
\label{fig:SHOT-VAE}
\end{figure}

We derive the following inequalities and use them to demonstrate $(3)$. 
\begin{equation}
    \begin{split}
        \vert \E_{\hat{p}(\rvy\vert \rmX)}\log \frac{p(\rvy)}{q_{\bm{\phi}}(\rvy\vert \rmX)} &-\E_{q_{\bm{\phi}}(\rvy\vert \rmX)}\log \frac{p(\rvy)}{q_{\bm{\phi}}(\rvy\vert \rmX)}\vert\\
        =\vert \sum_{i=1}^{K}(\text{smooth}(\1_{\rvy})_{i}-&\bm{\pi}_{\bm{\phi}}(\rmX)_{i})(\log p(\rvy)_i-\log \bm{\pi}_{\bm{\phi}}(\rmX)_{i})\vert\\
        \leq K\cdot \delta\cdot M+ &K\cdot \delta\cdot \sup_{i} \vert\log \bm{\pi}_{\bm{\phi}}(\rmX)_{i}\vert
    \end{split}
    \label{eq:margin}
\end{equation}
where $M$ is the upper bound of $\log p(\rvy)$, that is
\[\sup_{i}\vert\log p(\rvy)_{i}\vert\leq M.\]

Utilize the equation (\ref{eq:limitation}), we have
\begin{equation}
    \sup_{i} \vert \frac{\bm{\pi}_{\bm{\phi}}(\rmX)_{i}}{\text{smooth}(\1_{\rvy})_{i}}-1\vert \leq \delta\cdot \frac{K-1}{\epsilon}
    \label{eq:limitation2}
\end{equation}

When $\delta\rightarrow 0$, we have $\log(T\delta+1)=T\delta+\Delta(T\delta)$. Combining (\ref{eq:limitation2}), we have
\begin{equation}
    \sup_{i}\vert\log \bm{\pi}_{\bm{\phi}}(\rmX)_{i}\vert \leq \delta\cdot \frac{K-1}{\epsilon}  +\log \frac{1}{1-\epsilon}+\Delta(\delta)
    \label{eq:limitation3}
\end{equation}
Combining (\ref{eq:margin}) and (\ref{eq:limitation3}), we have 
\begin{equation}
    \begin{split}
        &\vert \E_{\hat{p}(\rvy\vert \rmX)}\log \frac{p(\rvy)}{q_{\bm{\phi}}(\rvy\vert \rmX)} -\E_{q_{\bm{\phi}}(\rvy\vert \rmX)}\log \frac{p(\rvy)}{q_{\bm{\phi}}(\rvy\vert \rmX)}\vert\\
        &\leq KM\delta+\frac{K(K-1)}{\epsilon}\delta^2+K\log \frac{1}{1-\epsilon}\delta+\Delta(\delta)
    \end{split}
\end{equation}
The bound $(7)$ states that, when $\delta\rightarrow 0$, the $\zeta$ can be arbitrarily small, which prove the limitation (\ref{eq:smooth-elbo-1}). 

\textbf{The convergence for degenerate distribution.} In original $\text{ELBO}_{\sD_{L}}$, the labels $\rvy$ are treated as latent variables directly, which equals to obey the empirical degenerate distribution, i.e. $\hat{p}(\rvy\vert \rmX)=1$. In this situation, $\epsilon\rightarrow 0$, and the second component in $(7)$ becomes
\[
\frac{K(K-1)}{\epsilon}\delta^2+K\log \frac{1}{1-\epsilon}\delta\rightarrow \infty
\]
which indicates that the limitation (\ref{eq:smooth-elbo-1}) may not converge under the degenerate distribution.
\subsection{Appendix B}
\label{Appendix.B}
\textbf{Proposition 2}  \textit{The inference result of discrete variable $\rvy$ satisfies the following inequality that $\forall i =1,\cdots,K$}
\[
    \begin{split}
        \vert \bm{\pi}_{\bm{\phi}}(\rmX)-\text{smooth}(\1_{\rvy})\vert_{i}
        \leq \sqrt{\frac{1}{2} \KL(\hat{p}(\rvy\vert \rmX) \Vert q_{\bm{\phi}}(\rvy\vert \rmX))}
    \end{split}
\]
\textbf{proof:}

The widely used \textbf{Pinsker's inequality} states that, if $P$ and $Q$ are two probability distributions on a measurable space $(\rmX,\Sigma)$, then 
\[
\delta(P,Q)\leq \sqrt{\frac{1}{2}\KL(P\Vert Q)}
\]
where 
\[
\delta(P,Q) =\sup\{\vert P(\rmA)-Q(\rmA)\vert\vert \rmA \in \Sigma \text{ is a measurable event.} \}
\]
In our situation, the discrete random variable $\rvy$ has the event set $\rmA\subset \Sigma=\{1,\ldots,K\}$, and the distribution $P,Q$ satisfies
\[
\vert P(\rmA)-Q(\rmA)\vert =\vert \sum_{i\in \rmA} p(\rvy=i)-q(\rvy=i)\vert 
\]
where $p(\rvy)=\text{Cat}(\rvy \vert \text{smooth}(\1_{\rvy}))$,  $q(\rvy)=\text{Cat}(\rvy \vert \bm{\pi}_{\bm{\phi}}(\rmX))$. For all $i=1,\ldots,K$, we have
\[
\vert P({i})-Q({i})\vert  \leq \delta(P,Q)
\]
and 
\[
\vert P({i})-Q({i}) \vert = \vert \bm{\pi}_{\bm{\phi}}(\rmX)-\text{smooth}(\1_{\rvy})\vert_{i}
\]
Then, with \textbf{Pinsker's inequality}, the proposition is easy to prove. 
\subsection{Appendix C}
\label{Appendix.C}
\textbf{Proposition 3} \textit{smooth-ELBO converges to $\text{ELBO}_{\sD_{L}}$ with the following equation}
\begin{equation*}
 \vert \text{smooth-ELBO}_{\sD_{L}}(\rmX,\rvy)-\text{ELBO}_{\sD_{L}}(\rmX,\rvy)\vert \leq C_1\delta+C_2\frac{\delta^2}{\epsilon}+\Delta(\delta)
\end{equation*}

As mentioned in $(7)$, we have

\begin{equation}
\begin{split}
         &\vert \text{smooth-ELBO}_{\sD_{L}}(\rmX,\rvy)-\text{ELBO}_{\sD_{L}}(\rmX,\rvy)\vert \\
         &= \vert \E_{\hat{p}(\rvy\vert \rmX)}\log \frac{p(\rvy)}{q_{\bm{\phi}}(\rvy\vert \rmX)} -\E_{q_{\bm{\phi}}(\rvy\vert \rmX)}\log \frac{p(\rvy)}{q_{\bm{\phi}}(\rvy\vert \rmX)}\vert\\
        &\leq KM\delta+\frac{K(K-1)}{\epsilon}\delta^2+K\log \frac{1}{1-\epsilon}\delta+\Delta(\delta)
\end{split}
\end{equation}
then $C_1=KM+K\log \frac{1}{1-\epsilon}$ and $C_2 = K(K-1)$.

Furthermore, in the original paper, we derive the  $\text{ELBO}_{\sD_{L}}(\rmX,\rvy)$ with the independent assumptions as well as the empirical degenerate distribution. \textbf{One problem is, whether the format also holds for other empirical estimation form of $\hat{p}(\rvy\vert\rmX)$, e.g., the smoothed $\hat{p}(\rvy\vert\rmX)$.} Utilizing the \textit{Jensen-Inequality}, we can derive the following equivalent form:
\begin{equation}
    \begin{split}
          & \log p(\rmX) = \log \E_{q_{\bm{\phi}}(\rvz\vert \rmX),\hat{p}(\rvy\vert \rmX)}\frac{p(\rmX,\rvz,\rvy)}{q_{\bm{\phi}}(\rvz\vert \rmX)\hat{p}(\rvy\vert \rmX)}\\
          &\geq \E_{q_{\bm{\phi}}(\rvz\vert \rmX),\hat{p}(\rvy\vert \rmX)} \log \frac{p(\rmX,\rvz,\rvy)}{q_{\bm{\phi}}(\rvz\vert \rmX)\hat{p}(\rvy\vert \rmX)} \\
        &= \E_{q_{\bm{\phi}},\hat{p}}\log p(\rmX\vert \rvz,\rvy) -\KL(q_{\bm{\phi}}(\rvz\vert \rmX)\Vert p(\rvz))\\
        &-\KL(\hat{p}(\rvy\vert \rmX) \Vert p(\rvy)) =\text{ELBO}_{\sD_{L}}(\rmX,\rvy).
    \end{split}
    \label{eq:smooth-elbo-2}
\end{equation}
which explicitly proves that $\text{ELBO}_{\sD_{L}}(\rmX,\rvy)$ holds for any reasonable empirical distribution form of $\hat{p}(\rvy\vert\rmX)$.

Combining the equation (\ref{eq:smooth-elbo-2}) and $(8)$, the convergence of \textbf{Proposition 3} is easy to prove.
\subsection{Appendix D}
\textbf{Proposition 4} \textit{The mixup strategy under the disentangled VAE models can be understood as calculating the \text{optimal interpolation} between two points $\rmX_{0},\rmX_{1}$ in input space with the maximum likelihood:}
\begin{equation}
\begin{split}
        \max_{\Tilde{X}} (1-\lambda)\cdot \log(p_{\bm{\theta}}(\Tilde{\rmX}\vert \rvz_0,\rvy_0))
        +\lambda\cdot \log(p_{\bm{\theta}}(\Tilde{\rmX}\vert \rvz_1,\rvy_1)).
\end{split}
\end{equation}

First, it is easy to prove that the mixup strategy $\Tilde{\rmX}=(1-\lambda)\rmX_{0}+\lambda\rmX_{1}$ can be understood as calculating the \text{optimal interpolation} between two points $\rmX_{0},\rmX_{1}$ in data space with the norm-2 distance:
\begin{equation}
   \min_{\Tilde{X}} (1-\lambda)\cdot \Vert\Tilde{\rmX}-\rmX_0\Vert_2^2+\lambda\cdot \Vert\Tilde{\rmX}-\rmX_1\Vert_2^2.
\end{equation}

In VAE models, with the generation process, we have 
\begin{equation}
    \rmX_0 = f_{\bm{\theta}}(\rvz_0,\rvy_0);\quad \rmX_1 = f_{\bm{\theta}}(\rvz_1,\rvy_1)
\end{equation}
then the distribution $p_{\bm{\theta}}(\Tilde{\rmX}\vert \rvz,\rvy)$ becomes
\begin{equation}
    p_{\bm{\theta}}(\Tilde{\rmX}\vert \rvz,\rvy)=C_{0}(\bm{\sigma})\cdot\exp{-\frac{\Vert \Tilde{\rmX}- f_{\bm{\theta}}(\rvz,\rvy)\Vert }{2\cdot C_{1}(\bm{\sigma})}}
\end{equation}
where $C_0,C_1$ are constants associated with the constant $\bm{\sigma}$.

Substituting $(11),(12)$ into $(9)$, we can obtain the equivalence of $(9)$ and $(10)$, which proves the proposition.
\subsection{Appendix E}
\subsubsection{E.1: The margin between $\log p(\rmX)$ and $\text{ELBO}(\rmX)$.}
\textbf{Proposition 5} \textit{The margin between the true log-likelihood $\log p(\rmX)$ and $\text{ELBO}(\rmX)$ under the independent assumptions is} 
\[
  \log p(\rmX)-\text{ELBO} = \KL(q_{\bm{\phi}}(\rvz\vert \rmX) q_{\bm{\phi}}(\rvy\vert \rmX)\Vert p(\rvz,\rvy \vert \rmX)
\]
\textbf{proof:}

\[
\begin{split}
    &\log p(\rmX)=\int_{\rvz,\rvy} q_{\bm{\phi}}(\rvz\vert \rmX) q_{\bm{\phi}}(\rvy\vert \rmX) \log p(\rmX)d \rvz d\rvy\\ &=\int_{\rvz,\rvy} q_{\bm{\phi}}(\rvz\vert \rmX) q_{\bm{\phi}}(\rvy\vert \rmX) \log \frac{p(\rmX,\rvz,\rvy)}{p(\rvz,\rvy\vert \rmX)}d \rvz d\rvy\\
    &=\E_{q_{\bm{\phi}}(\rvz\vert \rmX) q_{\bm{\phi}}(\rvy\vert \rmX)}\log \frac{p(\rmX,\rvz,\rvy)}{q_{\bm{\phi}}(\rvz\vert \rmX)q_{\bm{\phi}}(\rvy\vert \rmX)}\\
    &+\int_{\rvz,\rvy} q_{\bm{\phi}}(\rvz\vert \rmX) q_{\bm{\phi}}(\rvy\vert \rmX)\log \frac{q_{\bm{\phi}}(\rvz\vert \rmX) q_{\bm{\phi}}(\rvy\vert \rmX)}{p(\rvz,\rvy\vert \rmX)}\\
    &=\text{ELBO} + \KL(q_{\bm{\phi}}(\rvz\vert \rmX) q_{\bm{\phi}}(\rvy\vert \rmX)\Vert p(\rvz,\rvy \vert \rmX)
\end{split}
\]
\subsubsection{E.2: The optimal interpolation.}
\textbf{Proposition 6} \textit{The optimal interpolation derived from $\KL$ distance between $q_{\bm{\phi}}(\rvy\vert \bm{\pi}_{\bm{\phi}}(\rmX_0))$ and $q_{\bm{\phi}}(\rvy\vert \bm{\pi}_{\bm{\phi}}(\rmX_1))$  with $\lambda \in [0,1]$ can be written as}
\begin{equation*}
\begin{split}
        \min_{\Tilde{\bm{\pi}}} (1-\lambda)\cdot&\KL(\bm{\pi}_{\bm{\phi}}(\rmX_0)\Vert \Tilde{\bm{\pi}})+\lambda\cdot \KL(\bm{\pi}_{\bm{\phi}}(\rmX_1)\Vert \Tilde{\bm{\pi}})\\
        &\textbf{s.t. }\sum_{i=1}^{K}\Tilde{\bm{\pi}}_{i}=1 ;\Tilde{\bm{\pi}}_{i}\geq 0,
\end{split}
\end{equation*}
\textit{and the solution $\Tilde{\bm{\pi}}$ satisfying}
\begin{equation}
    \Tilde{\bm{\pi}} = (1-\lambda) \bm{\pi}_{\bm{\phi}}(\rmX_0) + \lambda\bm{\pi}_{\bm{\phi}}(\rmX_1).
\end{equation}
\textbf{proof}:

Denote $\bm{\pi}_{0}=\bm{\pi}_{\bm{\phi}}(\rmX_0)$ and $\bm{\pi}_{1}=\bm{\pi}_{\bm{\phi}}(\rmX_1)$. The Lagrange multiplier form of $(14)$ in \textbf{Proposition \romannumeral1} satisfies: 
\[
\mathcal{L}(\Tilde{\bm{\pi}},t) = (1-\lambda)\cdot\KL(\bm{\pi}_0\Vert \Tilde{\bm{\pi}})+\lambda\cdot \KL(\bm{\pi}_1\Vert \Tilde{\bm{\pi}})+t*(\sum_{i=1}^{K}\Tilde{\bm{\pi}}_{i}-1)
\]
and the related KKT conditions are 
\[
\begin{split} 
    \frac{\partial \mathcal{L}(\Tilde{\bm{\pi}},t)}{\partial \Tilde{\bm{\pi}}}&=t-\frac{(1-\lambda)\cdot \bm{\pi}_0 + \lambda\cdot \bm{\pi}_1}{\Tilde{\bm{\pi}}}=0\\
    &t*(\sum_{i=1}^{K}\Tilde{\bm{\pi}}_{i}-1)=0
\end{split}
\]
Solve the above equations, the closed form of $\Tilde{\bm{\pi}}$ is
\[
\Tilde{\bm{\pi}} = (1-\lambda)\cdot \bm{\pi}_0 + \lambda\cdot \bm{\pi}_1
\]
\subsection{Appendix F}
\subsubsection{F.1: The semi-supervised performance on MNIST and SVHN.}
We evaluate the inference accuracy of \textit{SHOT-VAE} with experiments on two benchmark datasets, MNIST and SVHN. In experiments, we consider five advanced VAE models as baselines, i.e. the standard VAE (M2)\citep{DBLP:conf/nips/KingmaMRW14}, stacked-VAE (M1+M2) \citep{DBLP:conf/nips/KingmaMRW14}, disentangled-VAE \citep{DBLP:conf/nips/NarayanaswamyPM17}, hyperspherical-VAE \citep{DBLP:conf/uai/DavidsonFCKT18}, and domain-VAE \citep{DBLP:conf/iclr/IlseTLW19}. For fairness, the backbones are all 4-layer MLPs with the same amount of parameters (approximately 1M) and the latent dimensions of $\rvz$ are 10 for MNIST and 32 for SVHN. The results presented in Table \ref{table:smooth-ELBO} show that our SHOT-VAE achieves competitive results to other VAE models without introducing additional domain knowledge or multi-stage structures.
\begin{table}[t]
\caption{Error rates for \textit{SHOT-VAE} and other advanced VAE models on MNIST with 100 labels and SVHN with 1000 labels.}
\label{table:smooth-ELBO}
\begin{center}
\begin{tabular}{ccc}
\toprule
Method & MNIST & SVHN \\
\midrule
M2  & $11.97(\pm 1.71)$ & $54.33(\pm 0.11)$ \\
M1+M2  & $3.33(\pm 0.14)$  & $36.02(\pm0.10)$   \\
Disentangled-VAE  & $9.71(\pm 0.91)$  & $38.91(\pm1.06)$ \\
Hyperspherical-VAE & $5.2(\pm 0.20)$ & /\ \\
Domain-VAE & $\mathbf{2.7}(\pm 1.30)$ & $32.17(\pm 1.20)$ \\
\textbf{smooth-ELBO}   & $3.14(\pm 0.19)$ & $\mathbf{29.38}(\pm 0.78)$     \\
\textbf{SHOT-VAE}   & $3.12(\pm 0.22)$ & $\mathbf{28.82}(\pm 0.49)$     \\
\bottomrule
\end{tabular}
\end{center}
\end{table}
\begin{table}[t]
\centering
\caption{p-value of ANOVA.}
\label{table:anova}
\begin{tabular}{ccc}
\toprule
 & CIFAR-10 & CIFAR-100 \\
\midrule
$\epsilon$        &   $0.47$       & $0.38$           \\
$\gamma$          &   $0.24$       & $0.27$       \\
\bottomrule
\end{tabular}
\end{table}
\subsubsection{F.2: Robustness analysis of hyper-parameters.}
We have introduced some hyper-parameters in training SHOT-VAE, which can be grouped into 2 categories: (1) Parameters to train a deep generative model, i.e. $\tau$ for the reparameterization tricks \cite{DBLP:conf/icml/RezendeMW14,DBLP:conf/iclr/JangGP17} and $\beta$ for the beta-VAE model \citep{DBLP:journals/corr/abs-1804-03599}. (2) Parameters to improve the semi-supervised learning performance, i.e. $\epsilon$ in smooth-label and $\gamma$ in the warm-up strategy. Following the previous works \cite{DBLP:conf/iclr/JangGP17,DBLP:conf/nips/Dupont18,DBLP:journals/corr/abs-1804-03599}, we set $\tau=0.67$ and $\beta =0.01$ to make generative models work.

For parameters related to semi-supervised performance, we also simply use the default value in previous works \cite{DBLP:journals/corr/abs-1804-03599,DBLP:conf/nips/MullerKH19}, setting $\epsilon = 0.001$ and $\gamma=5$. Here we use the statistic hypothesis testing method \textit{one-way ANOVA} \citep{tabachnick2007experimental} to test the null hypothesis of the above three hyper-parameters, that is, the semi-supervised learning performance is the same for different settings of parameters. For $\epsilon$, as stated in equation $(7)$, it should not be too large or too small. If $\epsilon$ is too large, the smoothed $\hat{p}(\rvy\vert \rmX)$ is over flexible and becomes too far from the basic degenerated distribution. If too small, then the convergence speed $\propto1/\epsilon$ may be too slow. Therefore, we set 5 value scales, i.e. $[10^{-5},10^{-4},10^{-3},10^{-2},10^{-1}]$. For $\gamma$, we also use 5 value scales, i.e. $[1,2,5,10,20]$. For each value, we performe 5 experiments with different random seeds to conduct ANOVA. The results in Table \ref{table:anova} do not reject the null hypothesis (all p-values $>0.1$), which proves that \textbf{the semi-supervised performance is robust to the selection of hyper-parameters}.
\begin{table}[t]
\centering
\caption{The results of combined SHOT-VAE.}
\label{table:combined-VAE}
\begin{tabular}{ccc}
\toprule
Method & CIFAR-10 & CIFAR-100 \\
\midrule
EnAET        &   $95.82$       & $77.08$           \\
FixMatch          &   $95.69$       & $76.82$       \\
ReMixMatch          &   $94.86$       & $74.82$     \\
UDA & $94.73$& $74.12$\\
SWSA & $95.00$&  $72.11$\\
Mean Teacher & $93.72$& $72.29$\\
SHOT-VAE + DA & $95.44\pm 0.27$& $75.91\pm 0.35$\\
SHOT-VAE + MT & $95.59\pm 0.24$& $76.79\pm 0.32$\\
\bottomrule
\end{tabular}
\end{table}
\subsection{Appendix G: Comparison with advanced SOTA methods in leaderboard.}
Recent works on semi-supervised learning can be grouped into 3 categories: (1) data augmentation based models (e.g., VAT). (2) consistency learning (e.g., Mean Teacher and $\Pi-$model). (3) generative models (e.g., GAN and VAE). As we know, combining different useful methods will improve the performance, and there are 5 combined methods in the leaderboard\footnote{\url{https://paperswithcode.com/sota/semi-supervised-image-classification-on-cifar}} that report better results than SHOT-VAE, i.e. EnAET \citep{DBLP:journals/corr/abs-1911-09265} combining generative models and data augmentation, FixMatch \citep{DBLP:journals/corr/abs-2001-07685}, ReMixMatch \citep{DBLP:journals/corr/abs-1911-09785} and UDA \citep{xie2019unsupervised} combining data augmentation and consistency learning, and SWSA \citep{DBLP:conf/iclr/AthiwaratkunFIW19} combining $\Pi-$model and fast-SWA. 

In our experiments, we choose the latest single models as baselines and claim that our model outperformed others. Moreover, our SHOT-VAE model can also easily combine other semi-supervised models and raise much better results. As shown in Table \ref{table:combined-VAE}, we combine SHOT-VAE with data augmentations (DA) and Mean Teacher (MT) separately and achieved competitive results in the leaderboard. 
\end{document}